\documentclass{article}
\usepackage{PRIMEarxiv}
\usepackage[utf8]{inputenc}
\usepackage[T1]{fontenc}
\usepackage{textcomp}
\usepackage{amsmath,amssymb,amsfonts}
\usepackage{hyperref,url,xurl}
\usepackage{booktabs,longtable,array,calc}
\usepackage{graphicx,grffile}
\usepackage{pdflscape,float}
\usepackage{microtype}
\usepackage{xcolor}
\newcommand{\hl}[1]{#1}
\usepackage[normalem]{ulem}
\providecommand{\ul}[1]{\uline{#1}}
\providecommand{\real}[1]{#1}

\hypersetup{hidelinks,hypertexnames=false}
\DeclareUnicodeCharacter{2212}{\ensuremath{-}}
\DeclareUnicodeCharacter{2010}{-}
\DeclareUnicodeCharacter{0131}{\i{}}
\DeclareUnicodeCharacter{0130}{\.{I}}
\DeclareUnicodeCharacter{015B}{\'{s}}

\title{VITAL: A framework for validating agentic AI in pharmaceutical drug development and clinical trial operations}
\date{}
\author{\begin{minipage}{0.97\textwidth}\centering
{\small Sandeep Burugupalli\textsuperscript{1}, Pedro Matos-Filipe\textsuperscript{2}, Andrea Hita Ardiaca\textsuperscript{2}, Adam Taylor\textsuperscript{3}, Alex Grigorenko\textsuperscript{1}, Natalia Jakubiak\textsuperscript{2}, Xin Yu\textsuperscript{4}, Raghav Mani\textsuperscript{4}, Sherry Mangla\textsuperscript{5}, Alvi Lai\textsuperscript{6}, Rachel C. Mercado*\textsuperscript{7}, Pratik Shah*\textsuperscript{8,9,10}, Raja Shankar\textsuperscript{6}}
\\[0.55em]
{\footnotesize *For correspondence: Rachel C. Mercado (\href{mailto:rachel.mercado@iqvia.com}{\nolinkurl{rachel.mercado@iqvia.com}}), Pratik Shah (\href{mailto:pratik.shah@uci.edu}{\nolinkurl{pratik.shah@uci.edu}})}
\\[0.5em]
{\footnotesize \textsuperscript{1}AstraZeneca Boston. 35 Gatehouse Drive, Waltham, MA 02451, United States \\
\textsuperscript{2}AstraZeneca Barcelona. Avinguda de Roma, 81, 08029 Barcelona, Spain \\
\textsuperscript{3}AstraZeneca Cambridge. Francis Crick Avenue Cambridge Biomedical Campus Cambridge, CB2 0AA, United Kingdom \\
\textsuperscript{4}NVIDIA. 2788 San Tomas Expy, Santa Clara, CA 95051, United States \\
\textsuperscript{5}IQVIA Frankfurt. Unterschweinstiege, 60549 Frankfurt am Main, Germany \\
\textsuperscript{6}IQVIA London. The Point, 37 North Wharf Road, London W2 1AF, United Kingdom \\
\textsuperscript{7}IQVIA Durham. 2400 Ellis Road, Durham, NC 27703, United States \\
\textsuperscript{8,9,10}Departments of Biomedical Engineering, Electrical Engineering \& Computer Science, Pathology \& Laboratory Medicine, University of California, Irvine, 501 Aldrich Hall, Irvine, CA 92697, United States}
\end{minipage}}
\begin{document}
\maketitle

\begin{abstract}
Current assurance approaches do not establish whether evidence is sufficient to validate, monitor and govern agentic artificial intelligence (AI) within a pharmaceutical context of use. Plausible outputs may conceal failures in retrieval, reasoning, orchestration or human-control points, where reference standards are unavailable. We developed and operationalised VITAL (Validation by Intended use, Traceability, Autonomy and Lifecycle evidence), a validation framework for agents in clinical development, through landscape analysis, regulatory and standards mapping, expert interviews and multi-stakeholder refinement. VITAL integrates context of use, controls, behavioural validation and lifecycle evidence within a risk-adjusted framework. We applied VITAL across epidemiology and treatment-landscape analysis, site selection, enrolment rescue and trial-design \hl{trade-off support} to translate autonomy, reversibility, transparency and patient impact into proportionate validation activities, evidence requirements and auditable reconstruction of system behaviour. VITAL may inform evidence requirements and lifecycle accountability for higher-risk pharmaceutical uses; evaluation should assess reproducibility across assessors, organisations and system configurations.

\end{abstract}

\keywords{Agentic AI; validation framework; clinical development; clinical operations; AI governance; pharmaceutical R\&D; implementation; drug development; regulatory standards}

\section{Introduction}\label{introduction}

Clinical development manages uncertainty through staged evidence generation across discovery, preclinical testing, protocolised clinical studies, regulatory review and post-market surveillance. Trial operations have largely been anchored in traditional clinical sites, whilst analytic judgement has remained predominantly human-led, episodic and constrained by predefined study designs. This architecture protects patient safety and evidentiary integrity, but it operates within increasingly data-intensive and time-sensitive portfolios in which faster access to analysis can affect developmental decisions\textsuperscript{1,2}.

Agentic AI changes the assurance problem by redistributing analytic agency across pharmaceutical research, clinical development and clinical-trial operations. Such systems have moved rapidly from conceptual architectures to early operational use in medicine, healthcare and the pharmaceutical industry. Recent literature describes systems capable of clinical workflow automation, collaborative diagnostic support and tool-mediated scientific or operational tasks\textsuperscript{3,4}, whilst the US Food and Drug Administration has deployed agentic capabilities to support regulatory and administrative workflows\textsuperscript{5}. However, evidence directly addressing agents in clinical development and clinical operations remains limited, with much of the field represented by early implementations. The combination of rapid emergence and limited validation may increase organisational risk\hl{, which arises} not only from model outputs but from interactions amongst data, prompts, orchestration, tool use, human review and downstream decisions\textsuperscript{6}. A plausible final output may conceal inappropriate retrieval, weak provenance, biased evidence, cumulative error caused by hallucination, non-determinism or ineffective human-review gates.

The assurance challenge is further intensified by the growing use of real-world data (RWD) and real-world evidence (RWE) across the drug development lifecycle\textsuperscript{1}, adaptive designs that incorporate accumulating evidence and decentralised trial elements, and digital health technologies that extend participation and measurement beyond conventional sites\textsuperscript{2,7,8}. These developments increase the volume, heterogeneity and time sensitivity of decision-relevant evidence to support more responsive trial conduct. However, acceleration also increases the burden on development teams to interpret the resulting \hl{evidence whose validity depends on data provenance}.

Regulatory and standards bodies establish expectations for risk-based assurance, transparency, human oversight, data governance, lifecycle monitoring, quality management, and documentation\textsuperscript{6,9,10}. These expectations provide a foundation for responsible AI use but do not specify how to demonstrate that a multi-step, tool-using system behaved acceptably within its authorised boundaries. In this manuscript, we define an agentic system as a software-based AI system that pursues human-defined objectives by planning, retrieving information, invoking tools, coordinating tasks and interacting with users or other systems across a workflow. This operational gap is becoming more consequential as healthcare systems move from narrow, task-specific models towards increasingly autonomous, goal-directed systems.

A recent study identified seven eligible healthcare studies, most of which were exploratory or simulation-based, where the authors highlighted substantial heterogeneity in system architectures and limited real-world clinical validation\textsuperscript{11}. Many healthcare agents remain at an early stage of development with no system yet demonstrating the robustness or autonomy required for clinical deployment\textsuperscript{11}. These findings reinforce the need for validation frameworks that assess final-output accuracy as well as autonomy boundaries, intermediate reasoning, inter-agent coordination, and lifecycle change.

Yet, current validation approaches provide only partial coverage for AI agents. Computer-system validation and computer-software assurance support risk-based testing, change control, and auditability, but were not designed for non-deterministic system execution pathways\textsuperscript{12}. Current AI/ML guidance from federal regulatory agencies addresses data quality, model evaluation, human factors and lifecycle monitoring but is less operationally specific about autonomy, tool permissions, prompting, orchestration, multi-agent hand-offs and intermediate traceability. Broader AI-governance frameworks address organisational controls and risk management without defining the evidence required to show acceptable agentic behaviour\textsuperscript{13,14}.

To address the aforementioned unmet needs, we developed and operationalised VITAL: a \hl{lifecycle-oriented, risk-based} validation framework \hl{built on system-level evidence} for agentic AI in pharmaceutical research, clinical development and clinical trial operations. Recent agent-focused approaches, including the Coalition for Health AI (CHAI)'s testing and evaluation framework\textsuperscript{15,16}, have strengthened the assessment of agent behaviour and reliability. VITAL is positioned downstream of these activities as it addresses how evidence generated through testing, reliability assessment and established software assurance should be interpreted for a specified pharmaceutical use. The relationship between VITAL and adjacent assurance approaches is further explored in the Discussion section and Supplementary Note 4. VITAL is derived from an original multi-method study integrating a targeted landscape analysis, regulatory and standards mapping, and primary qualitative evidence from subject-matter experts. The framework was refined through multi-stakeholder input spanning pharmaceutical clinical development, technology platform, and clinical-operations implementation perspectives. Furthermore, we applied VITAL across four pharmaceutical workflows: (1) epidemiological and treatment-landscape analysis, (2) site selection and prioritisation, (3) enrolment forecasting and rescue recommendation, and (4) trial design trade-off support. We aim to translate these evidence sources into an operational agentic AI validation method that treats the execution trajectory alongside the final output as a unit of validation and links autonomy, reversibility, transparency and patient consequence to proportionate controls. In doing so, VITAL operationalises agent-specific domains that existing software-assurance and AI-governance approaches do not address in detail, especially across the drug development lifecycle. 

\section{Results}\label{results}

\begin{figure}[!htbp]
\centering
\IfFileExists{media/figure1_vital_framework.pdf}{\includegraphics[width=\textwidth]{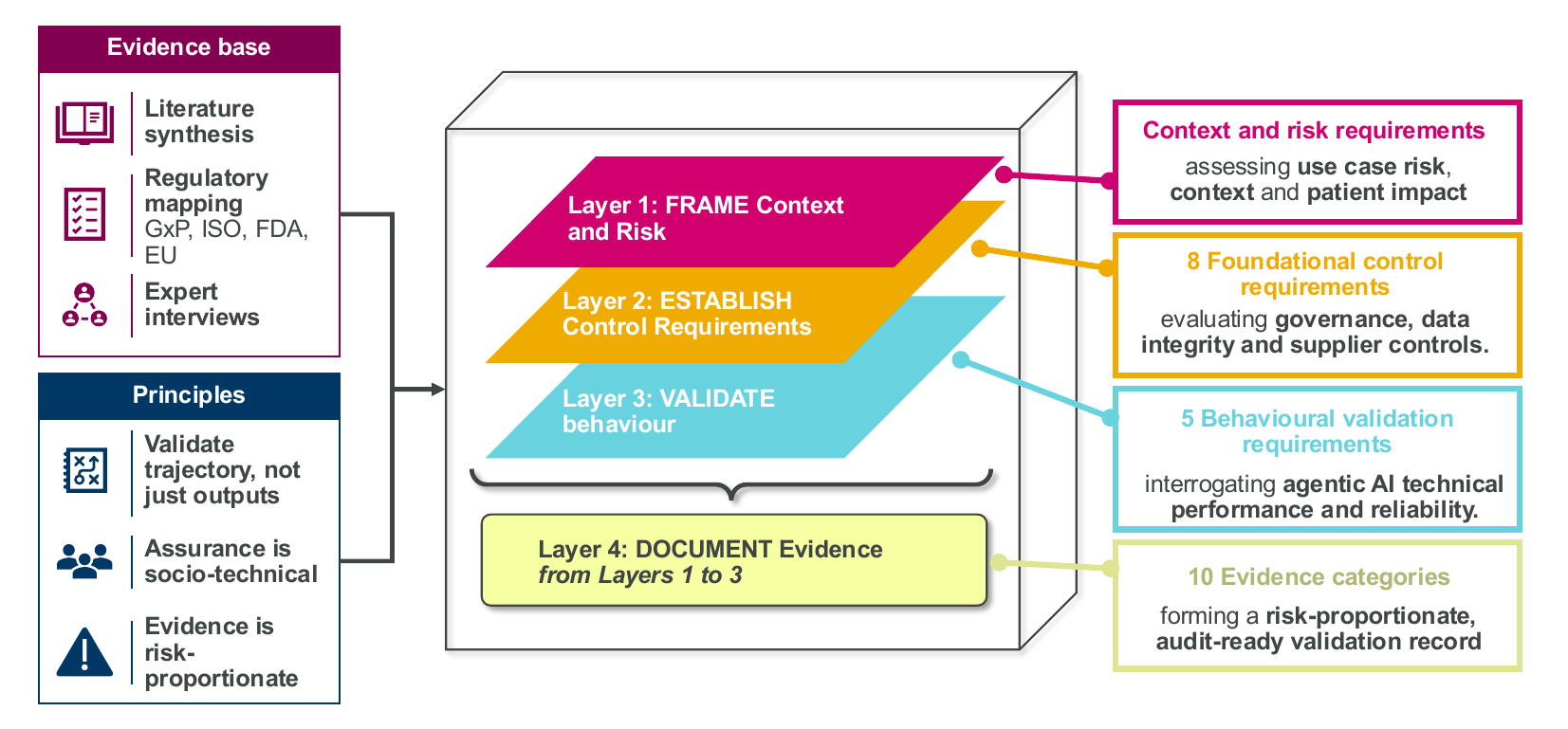}}{\fbox{\begin{minipage}[c][5cm][c]{0.83\linewidth}\centering\small Figure file not supplied.\\ media/figure1\_vital\_framework.pdf\end{minipage}}}
\caption{VITAL, a four-layer validation framework for agentic AI in clinical development. Documentation is proportionate to risk and traces evidence back to intended use, controls and agentic behavioural validation. See Table~\ref{tab:2} for the eight foundational control requirements, Table~\ref{tab:3} for the five behavioural validation domains, Table~\ref{tab:4} for the ten evidence categories.}
\label{fig:vital-framework}
\end{figure}

VITAL is a four-layer validation framework for agentic AI in drug development (Figure~\ref{fig:vital-framework}). The framework first defines context and risk, then establishes foundational controls, validates agentic behaviour, and documents proportionate evidence outputs. VITAL's central contribution is to shift validation from final outputs alone towards evaluation of agent behaviour, execution trajectories and lifecycle evidence, because a plausible yet incorrect final output can result from inappropriate intermediate steps, tool calls or hand-offs.

\subsection{Evidence base informing the framework}\label{evidence-base-informing-the-framework}

To inform the development of VITAL, we conducted a landscape analysis, regulations and standards mapping and analysis, expert interviews and refinement. These analyses provide the empirical and conceptual basis for the framework structure and underlying assumptions.

\subsubsection{Landscape analysis}\label{landscape-analysis}

A targeted landscape analysis of 113 peer-reviewed, regulatory, preprint and grey-literature sources identified recurring challenges associated with the validation of non-deterministic, tool-using and multi-agent systems. Five functional domains (reasoning and planning, tool use and execution, grounding and retrieval, multi-agent coordination and human-agent interaction) provided the initial organising structure for behavioural validation. Detailed methods, evidence composition and descriptive findings are reported in Supplementary Note 1 and Supplementary Tables~\ref{tab:s1}--\ref{tab:s3}.

\subsubsection{Standards and regulatory mapping}\label{standards-and-regulatory-mapping}

The standards and regulatory mapping identified a clear but incomplete baseline for agentic AI validation. The mapping covered four complementary domains: (1) regulated computerised systems and electronic records; (2) AI governance and risk management; (3) AI and ML development in pharmaceutical settings; and (4) clinical-trial or manufacturing quality assurance systems. Across GxP/CSV/Part 11\textsuperscript{17}; FDA CSA\textsuperscript{12} and GAMP 5\textsuperscript{29}; International Medical Device Regulators Forum (IMDRF) Good Machine Learning Practice (GMLP)\textsuperscript{18}; ISO/IEC 42001\textsuperscript{19} and ISO/IEC 23894\textsuperscript{20}; the EU AI Act\textsuperscript{21} and ICH E6(R3)\textsuperscript{22}; FDA/EMA Guiding principles of good AI Practice for drug development\textsuperscript{23}; and EU GMP Annex 22\textsuperscript{24} there was broad convergence on risk-based assurance, data integrity, lifecycle monitoring, change control, supplier oversight, and auditability.

These guidelines provide a foundation for agentic AI validation, consistent with current regulatory emphasis on context of use, risk-based evaluation, lifecycle management and proportionate quality oversight. However, the mapping also showed that existing guidance does not yet specify minimum requirements for constraining autonomy, governing tool permissions, reconstructing execution trajectories, securing multi-agent hand-offs, controlling orchestration changes or conducting scenario-based behavioural validation. VITAL therefore retains the standards-informed control baseline whilst providing greater operational specificity for agentic behaviour and its associated evidence outputs.

\begin{landscape}
\begin{figure}[H]
\centering
\IfFileExists{media/figure2_standards_crosswalk.pdf}{\includegraphics[width=\linewidth,height=0.82\textheight,keepaspectratio]{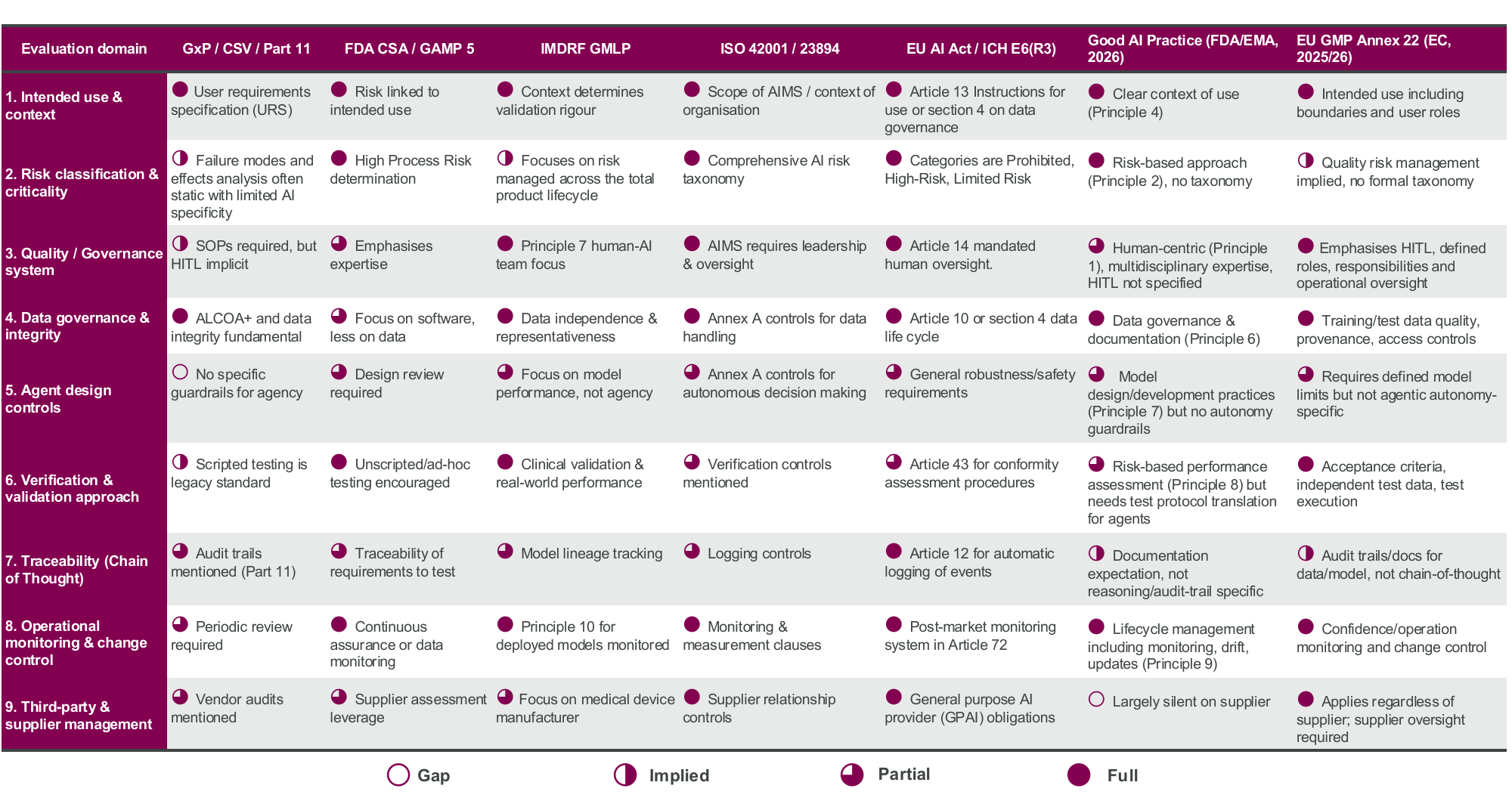}}{\fbox{\begin{minipage}[c][5cm][c]{0.83\linewidth}\centering\small Figure file not supplied.\\ media/figure2\_standards\_crosswalk.pdf\end{minipage}}}
\caption{Standards/regulatory mapping and analysis for agentic AI validation. Existing sources provide relatively strong coverage of foundational controls but weaker operational coverage of agent-specific behavioural validation.}
\label{fig:standards-crosswalk}
\end{figure}
\end{landscape}

Existing guidelines cited above provide relatively strong coverage for intended use, risk classification, quality/governance systems, data governance and integrity, and operational monitoring/change control (Figure~\ref{fig:standards-crosswalk}). These domains align with expectations in guidance for computerised-system validation, AI management systems, clinical trial quality management and AI systems for drug-development. For example, the FDA/EMA Guiding principles of good AI Practice for drug development\textsuperscript{23} support context-of-use definition to link validation effort to risk, data governance, monitoring deployed systems, and controlled system change.

By contrast, the analysis identified gaps and weaker operational coverage in domains that are central to agentic AI. Agent design controls were only partially addressed across most sources. Recent research has begun to address individual safeguards. For example, Task Shield constrains agent actions to those required by the user-defined task and evaluates this control against indirect prompt injection\textsuperscript{25}. GuardAgent translates specified safety policies into executable guardrail code and tests whether healthcare agents comply with privacy-related access controls\textsuperscript{26}. ToolSandbox evaluates stateful tool execution against intermediate and final workflow milestones\textsuperscript{27}, whilst MultiAgentBench assesses coordination across different multi-agent communication structures using milestone-based performance indicators\textsuperscript{28}. These methods address discrete technical safeguards but were developed as general-purpose research evaluations rather than validation approaches for regulated pharmaceutical workflows. They therefore do not specify how evidence should be assembled, governed and maintained for use by pharmaceutical sponsors, quality functions and regulators in approval, monitoring, inspection and risk-based revalidation across drug development. Traditional computer system validation (CSV) remains oriented towards predefined scripts and deterministic input-output expectations\textsuperscript{29}, whereas agentic systems require assessment across plausible trajectories, repeated runs, adversarial scenarios, tool-use pathways and human-agent interaction patterns (Figure~\ref{fig:standards-crosswalk}).

Traceability also emerged as a gap where our analyses (Figure~\ref{fig:standards-crosswalk}) found that whilst guidelines mentioned audit trails, event logging, documentation or model lineage, few define how to reconstruct the intermediate behaviour of an agent across planning, retrieval, tool invocation, hand-offs and output generation. For agentic AI, traceability could include the process by which outputs were produced, the sources accessed, the tools invoked, the decisions deferred to humans and the points at which guardrails or escalation rules were activated.

The standards analysis informed the validation framework in two ways. First, it established the minimum standards-informed coverage baseline for agentic AI validation. Second, it underpinned the framework's four-layer structure: context and risk should first be framed; foundational controls should then be established; agentic behaviour should be validated using evidence appropriate to autonomy and risk; and evidence outputs must be documented in a form suitable for oversight, audit and lifecycle management.

Taken together, existing sources establish a necessary control baseline but do not define the unit of validation or the evidence logic for agentic systems. VITAL adds three operational elements: explicit context-of-use and autonomy boundaries; trajectory-level validation of planning, retrieval, tool use, hand-offs and human-control points; and a risk-proportionate evidence package linked to monitoring and revalidation triggers.

\clearpage
\clearpage

\subsubsection{Expert perspectives}\label{expert-perspectives}

We interviewed 18 subject-matter experts (SMEs) from organisations operating in pharmaceutical development, life sciences, clinical-research services and AI computing. Participants held roles spanning clinical development, clinical operations, data science, product and platform engineering, infrastructure, enterprise security, governance and translational roles. Discussions considered applications including site identification, RWD cohort assembly, recruitment, monitoring, trial design, query management, enrolment forecasting and safety triage.

Our first key interview finding indicated that agentic systems cannot be sufficiently assessed through isolated output checking alone. Instead, validation was framed as an assessment of behaviour in context, including whether the system uses appropriate information, follows acceptable pathways, and remains within defined constraints. Participants additionally noted that correct-looking intermediate steps can still result in undesirable system-level behaviour when agents operate across tools, data sources or decision points. In site selection, a protocol-parsing component may correctly extract study requirements and a retrieval component may correctly obtain historical enrolment and RWD signals, however, an incomplete hand-off that omits current site capacity or data recency may still produce a misleading ranking. Such workflows increasingly combine multiple data sources and specialised components for feasibility assessment and site prioritisation\textsuperscript{30,31}. Similarly, an enrolment-rescue system may correctly identify deviation from plan but attribute it to site performance rather than eligibility criteria. In trial design, power calculations may be valid whilst the overall recommendation remains unsuitable because the orchestrator applies inappropriate assumptions or weights to eligibility, recruitment feasibility, participant burden or outcomes. Recent evaluation of LLM-generated trial designs found substantially lower agreement for eligibility criteria and outcome measures than for recruitment and intervention planning\textsuperscript{32}.

Our second finding was that oversight and evidentiary burden should scale with risk. Risk was described as a gradient influenced by agentic autonomy, reversibility, transparency and patient consequence with a distinction between lower-risk augmentation tasks and higher-stakes contexts affecting patients, workflows or organisational strategy.

Our third key finding indicated that layered evidence was needed instead of reliance on a single metric or benchmark. Validation requires a combination of retrospective comparison, human review, task-specific metrics, and ongoing monitoring. Some dimensions of performance are objective, whereas others are context dependent; validation, therefore, requires both measurement and judgement.

The SMEs additionally noted that observability was a prerequisite for trust, control, and post-hoc review. For example, telemetry, logging, and provenance were viewed as essential to enable auditability and the organisational ability to understand why a system produces a given recommendation and what evidence supports it.

SMEs also mentioned that agentic AI validation was contingent on workflow and data maturity. Agentic systems were seen as most suitable for processes that are structured, observable and governed, rather than for highly variable, poorly standardised or weakly controlled environments.

Finally, validation ownership was recommended to be distributed across technical, business and governance functions and be embedded from the outset in system design and architecture. Accountability for deployment and use cannot reside solely with technical teams; business ownership, governance oversight and cross-functional collaboration were all described as essential.

Areas of divergence were primarily differences in emphasis rather than disagreement with the need for agentic validation. Technical participants placed greater emphasis on architecture, tool permissions, telemetry and environment controls, whereas clinical, operational or governance participants more often emphasised workflow suitability, accountability and the evidence required for sign-off. These differences were treated as complementary requirements when refining the framework.

\textbf{\ul{VITAL framework}}

VITAL translates the agentic AI evidence base into four interdependent layers that define the context against which validation is judged, establish the controls required before testing, evaluate behaviours across trajectories, and preserve the resulting evidence for approval and lifecycle oversight (Figure~\ref{fig:vital-framework}). Detailed implementation examples of all four layers of the VITAL framework are outlined in the use case section.

\subsubsection{Layer 1: FRAME context and risk}\label{layer-1-frame-context-and-risk}

Layer 1 establishes the context against which all validation evidence is interpreted. The minimum output is a context-of-use and risk statement that defines intended users, workflow setting, objectives, permitted data sources, adjacent systems, decisions influenced, prohibited actions, escalation points and assumptions about human oversight.

Risk is treated as intrinsic use-case risk, assessed before controls are applied and then refined through residual-risk assessment. The core dimensions of risk are autonomy, reversibility, transparency/auditability and potential patient impact. Human-in-the-loop review is therefore specified as a control and validated as part of the workflow, rather than used as a substitute for defining intrinsic risk. Base agentic risk should reflect effective autonomy, i.e., the actions that the deployed architecture could plausibly reach through its models, tools, credentials, network connections and agent interfaces under foreseeable control failure or compromise rather than only actions permitted by the nominal workflow (nominal autonomy)\textsuperscript{33}. The resulting risk profile determines validation depth, documentation burden, monitoring intensity and revalidation triggers.

Consistent with established risk-based assurance principles, evidentiary burden should scale according to the agent's role in the workflow, degree of autonomy, reversibility, transparency and the potential consequence for patients, regulated records, trial conduct or organisational decision-making. To operationalise this principle, we defined a ten-level risk ladder for base agentic risk as the starting point for determining overall risk (Table~\ref{tab:1}). Unlike context-of-use profiles based principally on the downstream status of an output, the VITAL risk ladder also distinguishes the agentic capability through which that output is produced to separate advisory functions from other actions that have more autonomy (Table~\ref{tab:1}).

\begingroup\footnotesize
\setlength{\tabcolsep}{3pt}
\begin{longtable}[]{@{}
  >{\raggedright\arraybackslash}p{(\columnwidth - 8\tabcolsep) * \real{0.0990}}
  >{\raggedright\arraybackslash}p{(\columnwidth - 8\tabcolsep) * \real{0.1413}}
  >{\raggedright\arraybackslash}p{(\columnwidth - 8\tabcolsep) * \real{0.1369}}
  >{\raggedright\arraybackslash}p{(\columnwidth - 8\tabcolsep) * \real{0.2946}}
  >{\raggedright\arraybackslash}p{(\columnwidth - 8\tabcolsep) * \real{0.3282}}@{}}
\caption{VITAL ladder for assigning agentic AI base risk before contextual multipliers. Base risk increases with agent autonomy, workflow influence \hl{and} potential consequences, from information retrieval to adaptive autonomy. The ladder provides the starting risk estimate before applying contextual multipliers for reversibility, transparency and patient impact. Where a use case spans two adjacent autonomy definitions, a half-step may be assigned (e.g., 5.5 if a use case falls within advisory and advisory -- contextual).}\label{tab:1}\\
\toprule\noalign{}
\begin{minipage}[b]{\linewidth}\raggedright
\textbf{Base agentic risk (}\(\mathbf{A}\)\hl{\textbf{)}}
\end{minipage} & \begin{minipage}[b]{\linewidth}\raggedright
\textbf{Agent autonomy}
\end{minipage} & \begin{minipage}[b]{\linewidth}\raggedright
\textbf{Risk context}
\end{minipage} & \begin{minipage}[b]{\linewidth}\raggedright
\textbf{Risk definition}
\end{minipage} & \begin{minipage}[b]{\linewidth}\raggedright
\textbf{Examples}
\end{minipage} \\
\midrule\noalign{}
\endfirsthead
\toprule\noalign{}
\begin{minipage}[b]{\linewidth}\raggedright
\textbf{Base agentic risk (}\(\mathbf{A}\)\hl{\textbf{)}}
\end{minipage} & \begin{minipage}[b]{\linewidth}\raggedright
\textbf{Agent autonomy}
\end{minipage} & \begin{minipage}[b]{\linewidth}\raggedright
\textbf{Risk context}
\end{minipage} & \begin{minipage}[b]{\linewidth}\raggedright
\textbf{Risk definition}
\end{minipage} & \begin{minipage}[b]{\linewidth}\raggedright
\textbf{Examples}
\end{minipage} \\
\midrule\noalign{}
\endhead
\bottomrule\noalign{}
\endlastfoot
\(\mathbf{A}\)\textbf{1} & Informational & \textbf{Retrieve} & Static information access, no synthesis, judgement or prioritisation & Search protocol text, retrieve information \\
\(\mathbf{A}\)\textbf{2} & Informational & \textbf{Summarise} & Synthesis or formatting without evaluative judgement & Summarise protocol operational requirements, extract visit schedule, synthesise interpretation \\
\(\mathbf{A}\)\textbf{3} & Informational to analytical & \textbf{Surface signals} & Descriptive patterns or trends without prioritisation or recommendations & Surface descriptive protocol burden indicators (intensive visit schedule, many procedures/assessments, tight visit windows) \\
\(\mathbf{A}\)\textbf{4} & Analytical & \textbf{Apply rules} & Deterministic rule-based scoring or classification, transparent thresholds or rubrics & Burden scoring rubric, classify feasibility risk via fixed thresholds \\
\(\mathbf{A}\)\textbf{5} & Advisory & \textbf{Recommend} & Advisory decision support, human decides, outputs non-binding, explainable rationale required & Recommend feasibility mitigations, rank endpoint candidates \\
\(\mathbf{A}\)\textbf{6} & Advisory (contextual) & \textbf{Recommend with context} & Context-aware advisory, adapts recommendations using multi-factor signals, still non-binding & Recommend and rank sites using contextual operational signals (geography, start-up timelines, staffing/capacity, competing trials) and propose enrolment rescue options \\
\(\mathbf{A}\)\textbf{7} & Autonomous execution & \textbf{Execute (bounded)} & Executes pre-approved actions within scope & Auto-send site follow-ups, auto-create tickets/tasks \\
\(\mathbf{A}\)\textbf{8} & Goal-directed autonomy & \textbf{Act to achieve goal} & Plans how to achieve a specific objective and executes steps, higher non-determinism & Plan and execute a bounded workflow to close monitoring findings (retrieve open findings \textgreater{} draft responses \textgreater{} route for review \textgreater{} track closure) \\
\(\mathbf{A}\)\textbf{9} & Optimising autonomy & \textbf{Multi-objective trade-offs} & Balances competing objectives, implicit value judgements, harder to explain or contest & Optimise monitoring allocation across cost/speed/risk, enrolment strategy balancing quality vs velocity \\
\(\mathbf{A}\)\textbf{10} & Adaptive autonomy & \textbf{Change behaviour over time} & Learns \& updates decision logic, lifecycle validation required & Self-updating feasibility scoring, continuous learning \\
\end{longtable}
\endgroup

Once the base agentic risk is identified, the overall risk is estimated by combining the base agentic risk score \(\mathbf{(A)}\) with multipliers for reversibility (\(\mathbf{M}_{\mathbf{reversibility}}\)), transparency (\(\mathbf{M}_{\mathbf{transparency}}\)), and patient impact (\(\mathbf{M}_{\mathbf{patient}}\)). Reversibility adjusts risk according to the ease with which an agentic output or action can be detected, corrected or undone. Read-only advisory or easily reversible outputs attract a lower multiplier, whereas actions that are difficult to reverse, propagate across systems or influence downstream operational decisions require a higher multiplier (Equation 1). Transparency adjusts risk according to the extent to which users, reviewers\hl{ and} auditors can understand how an agent generated its output. Higher transparency requires reconstructable trajectories, identifiable sources, and reviewable tool calls; lower transparency increases validation burden because errors are harder to detect, challenge, or explain. The patient-impact multiplier adjusts risk according to the plausible consequence of agent behaviour for participant eligibility, trial conduct, safety processes, treatment-relevant decisions, or other patient-adjacent outcomes. Where direct patient care, safety-critical action or patient eligibility is affected, the patient-impact multiplier should dominate the final classification (Equation 2). For broader applications within pharmaceutical organisations, the patient-impact multiplier could be replaced with a domain-consequence multiplier to reflect the plausible effects of agent behaviour on other regulated or business-critical outcomes such as the integrity of financial reporting and associated internal controls in processes subject to the Sarbanes-Oxley Act\textsuperscript{34}. The multipliers are illustrative and are intended to operationalise the framework's risk logic. The values were chosen to create sufficient spread across risk categories to demonstrate materially different risk conditions that lead to different evidence requirements. Organisations can calibrate thresholds and multipliers against their governance model, risk-appetite and use-case portfolio. Transfer to another regulated domain would require re-specifying the consequence dimension and calibration thresholds for that domain.

Equation 1: Risk-adjusted classification formula for agentic AI validation. Multipliers for reversibility, transparency and patient impact are illustrative.

\[\mathbf{R = A\  \times \ }\mathbf{M}_{\mathbf{reversibility}}\mathbf{\  \times \ }\mathbf{M}_{\mathbf{transparency}}\mathbf{\  \times \ }\mathbf{M}_{\mathbf{patient}}\]

Where:

\begin{itemize}
\item
  \(\mathbf{R =}\) overall risk score of the agentic system
\item
  \(\mathbf{A =}\) base agentic risk

  \begin{itemize}
  \item
    \(\mathbf{A}\) \textbf{1 − 3} = retrieve/summarise/surface signals
  \item
    \(\mathbf{A}\) \textbf{4} = apply rules
  \item
    \(\mathbf{A}\) \textbf{5 − 6} = recommend/recommend with context
  \item
    \(\mathbf{A}\) \textbf{7} = execute bounded actions
  \item
    \(\mathbf{A}\) \textbf{8 − 10} = goal-directed, \hl{optimising}, or adaptive autonomy
  \end{itemize}
\item
  \(\mathbf{M}_{\mathbf{reversibility}}\) = reversibility multiplier
\item
  \(\mathbf{M}_{\mathbf{transparency}}\) = transparency multiplier
\item
  \(\mathbf{M}_{\mathbf{patient}}\) = patient-risk multiplier
\end{itemize}

\[\mathbf{M}_{\mathbf{reversibility}}\mathbf{\  \in \ }\left\{ \mathbf{1}\mathbf{.}\mathbf{0}\mathbf{,\ }\mathbf{1}\mathbf{.}\mathbf{25}\mathbf{,\ }\mathbf{1}\mathbf{.}\mathbf{5} \right\}\]

1.0 = easy to reverse

1.25 = partially reversible

1.5 = difficult to reverse

\[\mathbf{M}_{\mathbf{transparency}}\mathbf{\  \in \ }\left\{ \mathbf{1}\mathbf{.}\mathbf{0}\mathbf{,\ }\mathbf{1}\mathbf{.}\mathbf{25}\mathbf{,\ }\mathbf{1}\mathbf{.}\mathbf{5} \right\}\]

1.0 = transparent

1.25 = partly explainable

1.5 = lower transparency, harder to contest

\[\mathbf{M}_{\mathbf{patient}}\mathbf{\  \in \ }\left\{ \mathbf{1}\mathbf{.}\mathbf{0}\mathbf{,\ }\mathbf{1}\mathbf{.}\mathbf{25}\mathbf{,\ }\mathbf{1}\mathbf{.}\mathbf{5}\mathbf{,\ }\mathbf{3} \right\}\]

1.0 = no plausible patient impact

1.25 = indirect patient impact

1.5 = patient eligibility implications

3.0 = direct patient care, safety-critical action

Equation 2: Interpretation of risk-adjusted validation classification. Risk-adjusted scores map agentic systems into low, moderate, high, very high or critical validation categories. These categories determine the minimum evidence requirement across VITAL, including context-of-use specification, foundational controls, behavioural validation, traceability, monitoring and revalidation requirements. Patient-facing or patient-adjacent use cases should not be classified solely by numerical score. Minimum floor classifications should be applied to prevent underestimation of risk arising from apparently low autonomy or high human oversight.

\textbf{Risk score interpretation:}

\begin{itemize}
\item
  Low: \(R\  < \ 5\)
\item
  Moderate: \(5\  \leq \ R\  < 10\)
\item
  High: \(10\  \leq \ R\  < 20\)
\item
  \hl{Very High}: \(20\  \leq \ R\  < 35\)
\item
  Critical: \(R\  \geq 35\)
\end{itemize}

\textbf{Patient floor rule:}

\begin{itemize}
\item
  If \(\mathbf{M}_{\mathbf{patient}}\mathbf{= 1.5}\), then use case should be classified \textbf{High or above}
\item
  If \(\mathbf{M}_{\mathbf{patient}}\mathbf{= 3}\), then use case should be classified \textbf{Very High or above}
\item
  If \hl{\(\mathbf{M}_{\mathbf{patient}}\mathbf{= 3\ \text{and}\ A\  \geq 7}\)}, then classify as \textbf{Critical}
\end{itemize}

\subsubsection{Layer 2: ESTABLISH foundational control requirements}\label{layer-2-establish-foundational-control-requirements}

Layer 2 defines the minimum control baseline required before agentic behavioural validation. The eight requirements cover \hl{quality and governance system, data governance and integrity}, agent design, verification and validation, traceability, operational monitoring and change control, \hl{third-party and supplier management} and transparency and auditability. Examples include least-privilege tool permissions and approved source registers (Table~\ref{tab:2}). These controls establish the governed environment in which validation evidence can be generated, interpreted and audited.

\begingroup\footnotesize
\setlength{\tabcolsep}{3pt}
\begin{longtable}[]{@{}
  >{\raggedright\arraybackslash}p{(\columnwidth - 2\tabcolsep) * \real{0.2350}}
  >{\raggedright\arraybackslash}p{(\columnwidth - 2\tabcolsep) * \real{0.7650}}@{}}
\caption{Eight foundational control requirements for agentic AI systems.}\label{tab:2}\\
\toprule\noalign{}
\begin{minipage}[b]{\linewidth}\raggedright
\textbf{Eight control requirements}
\end{minipage} & \begin{minipage}[b]{\linewidth}\raggedright
\textbf{Description}
\end{minipage} \\
\midrule\noalign{}
\endfirsthead
\toprule\noalign{}
\begin{minipage}[b]{\linewidth}\raggedright
\textbf{Eight control requirements}
\end{minipage} & \begin{minipage}[b]{\linewidth}\raggedright
\textbf{Description}
\end{minipage} \\
\midrule\noalign{}
\endhead
\bottomrule\noalign{}
\endlastfoot
Quality and governance system & The organisational structures, roles, responsibilities, procedures and escalation pathways through which the system is designed, approved, deployed and overseen over its lifecycle. \\
Data governance and integrity & The controls and processes ensuring the provenance, quality, representativeness, privacy, access management, and lifecycle integrity of the data used for prompting, retrieval, system outputs and evaluation. \\
Agent design controls & The independently enforced technical and procedural controls that bound effective autonomy, including least-privilege permissions, tool and network access, credential isolation, blocked actions, stopping rules, escalation conditions and fail-safe interruption. \\
Verification and validation approach & The structured testing strategy used to demonstrate that the system is built as intended (verification) and performs acceptably within its intended context of use (validation), including methods, test design and acceptance criteria. \\
Traceability of system behaviour & The ability to reconstruct system behaviour after the fact, including logging of inputs, retrieved sources, tool calls, outputs, versions and decision traces. \\
Operational monitoring and change control & The processes used to monitor system performance during operational use, detect degradation or drift and manage changes, including criteria and triggers for revalidation. \\
Third-party and supplier management & The identification and oversight of external models, tools, APIs, vendors and data sources, evaluation infrastructure and runtime dependencies, including ownership, integrity verification, incident handling and contingency arrangements. \\
Transparency and auditability & The mechanisms that enable an independent reviewer to understand system behaviour, including what actions were taken, why those actions occurred and what evidence was available at a specific point in time. \\
\end{longtable}
\endgroup

\subsubsection{Layer 3: VALIDATE behaviour}\label{layer-3-validate-behaviour}

Layer 3 evaluates whether the agentic system behaves acceptably within its specified context of use. Because agentic failures may arise from intermediate reasoning, retrieval, tool use or hand-offs, validation can include trajectory review, scenario-based testing, adversarial or simulation-based testing, benchmark or historical comparisons where available, and structured subject-matter expert adjudication where gold standards do not exist (Table~\ref{tab:3}).

\begingroup\footnotesize
\setlength{\tabcolsep}{3pt}
\begin{longtable}[]{@{}
  >{\raggedright\arraybackslash}p{(\columnwidth - 2\tabcolsep) * \real{0.2986}}
  >{\raggedright\arraybackslash}p{(\columnwidth - 2\tabcolsep) * \real{0.7014}}@{}}
\caption{Five behavioural validation domains for agentic AI systems.}\label{tab:3}\\
\toprule\noalign{}
\begin{minipage}[b]{\linewidth}\raggedright
\textbf{Five behavioural validation domains}
\end{minipage} & \begin{minipage}[b]{\linewidth}\raggedright
\textbf{Description}
\end{minipage} \\
\midrule\noalign{}
\endfirsthead
\toprule\noalign{}
\begin{minipage}[b]{\linewidth}\raggedright
\textbf{Five behavioural validation domains}
\end{minipage} & \begin{minipage}[b]{\linewidth}\raggedright
\textbf{Description}
\end{minipage} \\
\midrule\noalign{}
\endhead
\bottomrule\noalign{}
\endlastfoot
Reasoning and planning & Ability of the system to decompose objectives into coherent and logically ordered steps. \\
Tool use and execution & The correctness, safety, and appropriateness of interaction with external systems, including whether tool calls remain authorised and whether the agent attempts to expand privileges, evade restrictions or exploit its execution environment. \\
Retrieval augmented generation (RAG) & Whether the system retrieves, synthesises and cites relevant information from authorised sources whilst resisting indirect prompt injection and unsupported claims. \\
Single and multi-agent coordination & Assessment of single-agent performance and the integrity of inter-agent hand-offs, including message provenance, authentication, trust boundaries, replay protection and containment of a compromised component. \\
Human-agent interaction & Evaluation of human oversight and interpretability, including whether approval, pause and termination controls remain effective across extended trajectories. \\
\end{longtable}
\endgroup

The effectiveness of human oversight depends on the presence of a reviewer and on the point at which the reviewer enters the execution trajectory and \hl{on} the information available at that point. A downstream reviewer may be able to reject an implausible final recommendation but remain unable to detect an earlier retrieval omission, corrupted hand-off or inappropriate tool invocation if those events are not exposed in the review interface. Conversely, approval gates positioned too early may provide little control over subsequently generated actions. Validation of human-agent interaction should therefore determine whether failure states remain observable and contestable at the point where human intervention is expected to occur. Oversight should be regarded as effective only where the reviewer has sufficient information, authority and temporal opportunity to prevent or contain the relevant consequence.

\subsubsection{Layer 4: DOCUMENT evidence outputs}\label{layer-4-document-evidence-outputs}

Layer 4 converts the preceding layers (Layers 1-3) into an auditable validation record proportionate to risk by documenting intended use, controls, test evidence, deviations, monitoring and revalidation logic in a format that is reviewable by governance, quality, technical and business stakeholders. The risk profile (as described in Layer 1) dictates the level of documentation required. Representative outputs include the context-of-use specification, validation and revalidation protocol, traceability and logging specification, operational monitoring plan and third-party dependency register (Table~\ref{tab:4}).

\begingroup\footnotesize
\setlength{\tabcolsep}{3pt}
\begin{longtable}[]{@{}
  >{\raggedright\arraybackslash}p{(\columnwidth - 2\tabcolsep) * \real{0.3080}}
  >{\raggedright\arraybackslash}p{(\columnwidth - 2\tabcolsep) * \real{0.6920}}@{}}
\caption{Ten evidence categories for agentic AI system validation.}\label{tab:4}\\
\toprule\noalign{}
\begin{minipage}[b]{\linewidth}\raggedright
\textbf{Ten evidence categories}
\end{minipage} & \begin{minipage}[b]{\linewidth}\raggedright
\textbf{Description}
\end{minipage} \\
\midrule\noalign{}
\endfirsthead
\toprule\noalign{}
\begin{minipage}[b]{\linewidth}\raggedright
\textbf{Ten evidence categories}
\end{minipage} & \begin{minipage}[b]{\linewidth}\raggedright
\textbf{Description}
\end{minipage} \\
\midrule\noalign{}
\endhead
\bottomrule\noalign{}
\endlastfoot
Context-of-use specification & Describes the intended use, user population, workflow context, permitted data sources, decisions, and system boundaries, providing a clear statement against which validation is assessed. \\
Governance operating model & Defines accountability for design, approval, deployment, exception handling and oversight, including the decision gates, escalation pathways, and the allocation of human-in-the-loop or human-on-the-loop control responsibilities. \\
Data specification & Demonstrates the provenance, quality, representativeness, access controls and lifecycle handling of the data used for prompts, retrieval, outputs, and evaluation with sufficient detail for an independent reviewer to assess fitness for purpose. \\
Design specifications and controls & Describes the system architecture and behavioural constraints, including agent roles, tool access, permissions, blocked actions, escalation triggers, and the guardrails used to prevent unsafe or unauthorised behaviour. \\
Validation and revalidation protocol & Sets out the validation test strategy, including test methods (e.g., retrospective checks, benchmarking against reference datasets, simulation or adversarial testing, subject matter expert review), acceptance criteria, and the rules for pass/fail decisions, deviations, and revalidation triggers. \\
Validation evidence package & Contains the structured set of outputs generated from validation activities, including test results, performance summaries, expert assessments, deviation records, and conclusions that demonstrate that the system meets predefined acceptance criteria for its intended use. \\
Traceability and logging specification & Defines the information required to reconstruct system behaviour, including logging and retention of inputs, retrieved sources, intermediate steps, tool calls, outputs, system versions, and decision traces. \\
Operational monitoring plan & Specifies the approach to monitoring system performance in production, including metrics, alert thresholds, detection of drift or degradation, triggers for revalidation, and controls governing system updates and release management. \\
Third-party dependency register & Identifies all external dependencies, including models, APIs, vendors, and data sources, defining ownership, risk management responsibilities, incident handling, and contingency or fallback arrangements. \\
Transparency and auditability specification & Documents how system behaviour is made observable and reviewable, including mechanisms for exposing decision logic, evidence sources, and execution traces, enabling independent reviewers to understand, verify, and assess the system's actions at a given point in time. \\
\end{longtable}
\endgroup

To distinguish the underlying evidence package from the rationale by which that evidence is translated into a governance disposition, a minimum record for reconstruction of the validation decision is provided in Supplementary Note 3 and Supplementary Table~\ref{tab:s5}.

\begin{figure}[H]
\centering
\IfFileExists{media/figure3_decision_tree.pdf}{\includegraphics[width=0.95\textwidth,height=0.82\textheight,keepaspectratio]{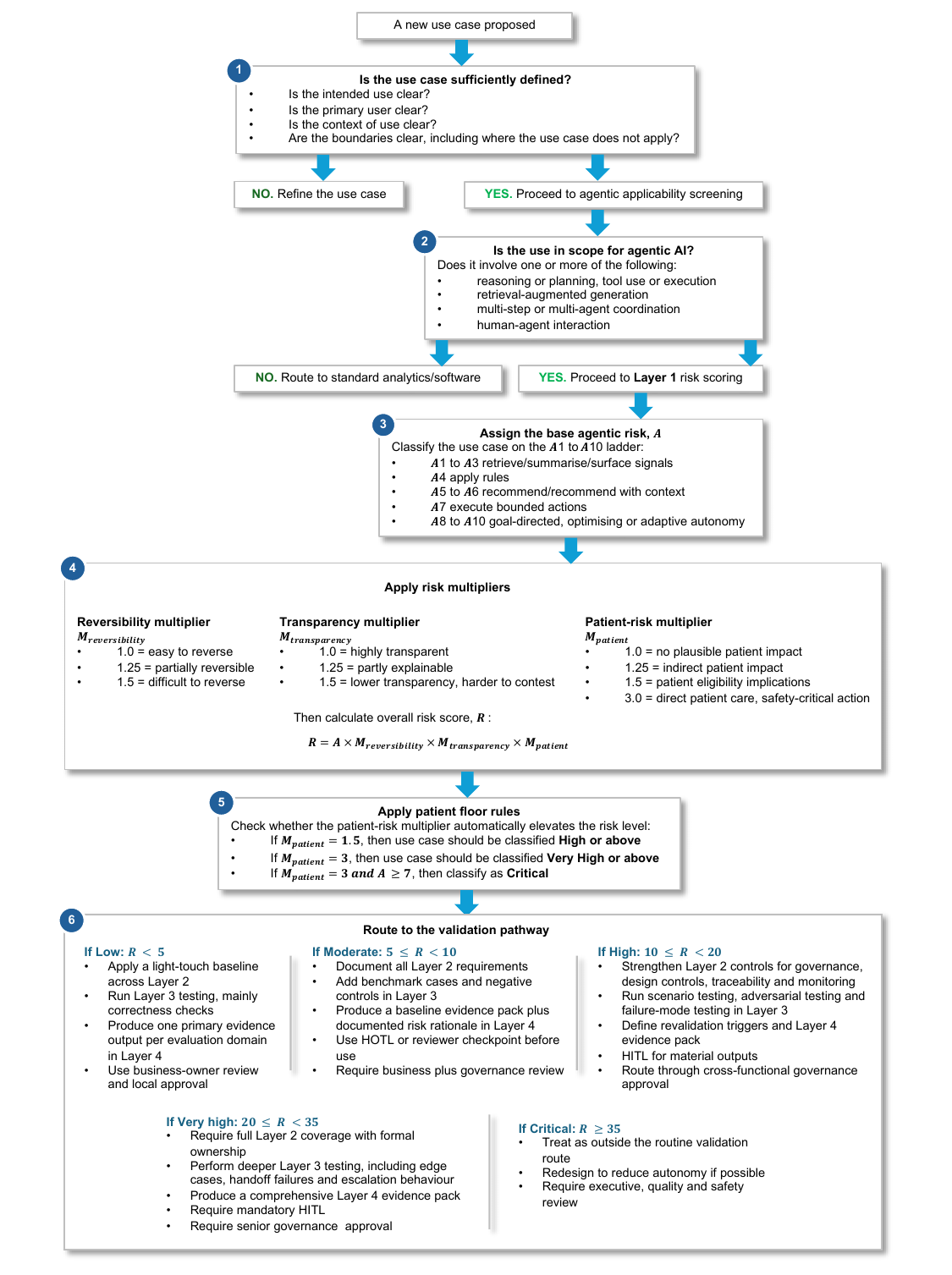}}{\fbox{\begin{minipage}[c][5cm][c]{0.83\linewidth}\centering\small Figure file not supplied.\\ media/figure3\_decision\_tree.pdf\end{minipage}}}
\caption{VITAL decision tree for risk-proportionate validation. The decision tree converts intended use, agent autonomy, reversibility, transparency, patient impact and regulated-workflow relevance into a validation category and minimum evidence package for each of the five risk-adjusted tiers (low, moderate, high, very high, critical).}
\label{fig:decision-tree}
\end{figure}

To support application, all four VITAL layers are operationalised through an end-to-end decision tree that links intended use, base agentic risk, contextual risk multipliers and evidence outputs designed to guide users from initial use-case framing to risk-adjusted validation requirements (Figure~\ref{fig:decision-tree}).

\clearpage
\clearpage

\subsection{Application across four clinical-development use cases}\label{application-across-four-clinical-development-use-cases}

We applied VITAL to four representative clinical-development use cases selected to span different lifecycle stages and forms of agentic AI functionality. The applications provide structured demonstration of how context of use, autonomy, reversibility, transparency and patient impact translate into validation requirements. The use cases operate upstream of protocol design, during study start-up, during trial conduct and during design and planning, and range from goal-directed analysis to context-aware recommendation and multi-objective optimisation. For each use case, base agentic risk was assigned before controls were considered. The resulting pre-control risk category was used to specify validation depth, evidence requirements, monitoring intensity and revalidation triggers. Residual risk can subsequently be assessed after application of controls such as human review, restricted tool permissions, traceability, access management and change control.

\subsubsection{Epidemiology and treatment-landscape analysis}\label{epidemiology-and-treatment-landscape-analysis}

\begin{landscape}
\begin{figure}[H]
\centering
\IfFileExists{media/figure4_use_case_cards.pdf}{\includegraphics[width=\linewidth,height=0.82\textheight,keepaspectratio]{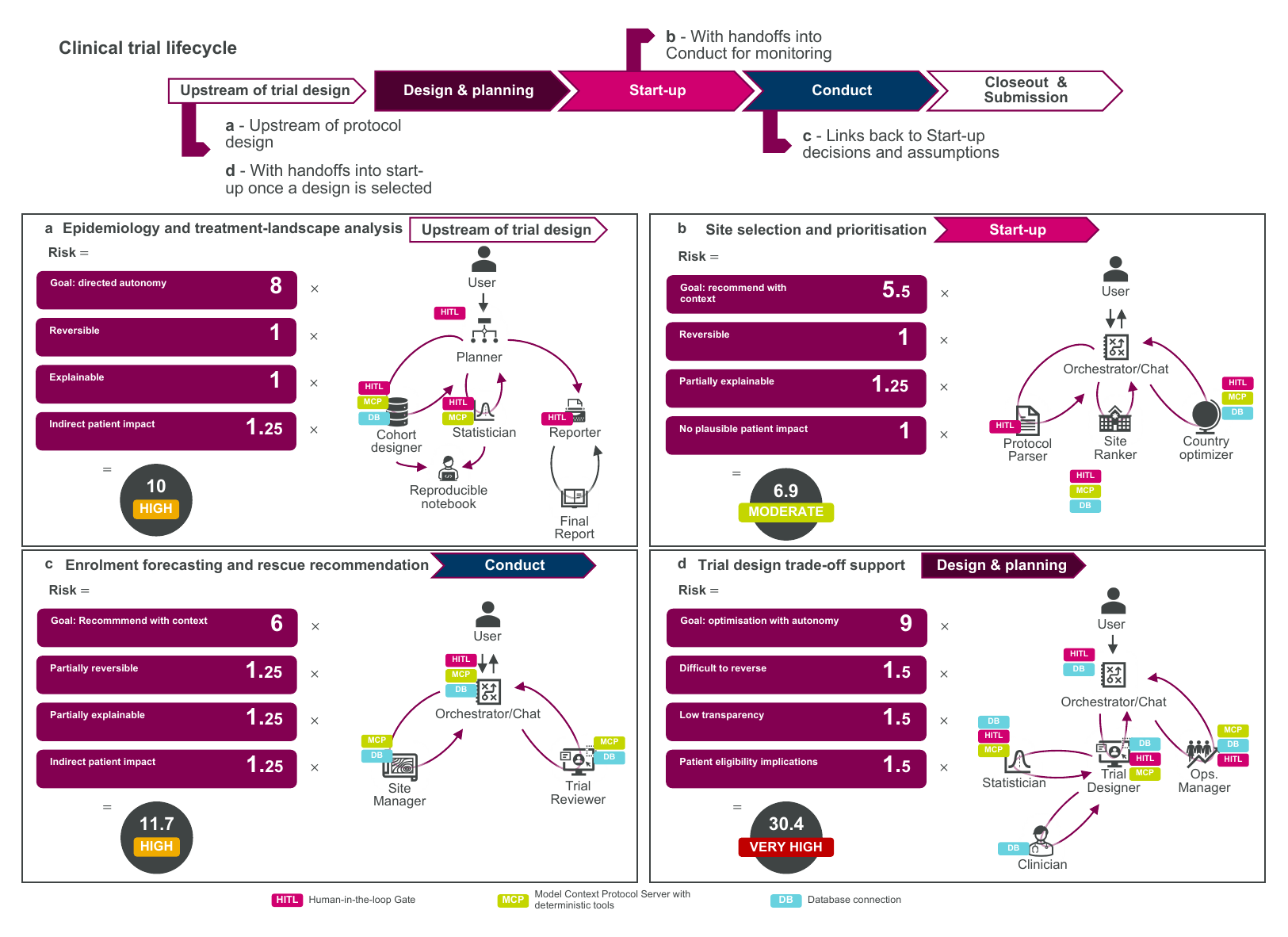}}{\fbox{\begin{minipage}[c][5cm][c]{0.83\linewidth}\centering\small Figure file not supplied.\\ media/figure4\_use\_case\_cards.pdf\end{minipage}}}
\caption{Application of VITAL to four representative clinical-development use cases (a) Epidemiology and treatment-landscape analysis. (b) Site selection and prioritisation. (c) Enrolment forecasting and rescue recommendation. (d) Trial design trade-off support. The risk scores are illustrative and can be modified according to organisational considerations.}
\label{fig:use-case-cards}
\end{figure}
\end{landscape}

Our decision tree for VITAL (Figure~\ref{fig:decision-tree}) can be applied prospectively when a new use case is proposed for validation. For the epidemiology and treatment-landscape use case, which operates upstream of protocol design, the intake would begin by defining the proposed objective, authorised data sources, analytical tools, human-approval points and prohibited downstream actions. The proposed system would then be classified as agentic because it decomposes a research question and coordinates a multi-step analytical workflow across components and tools. A planner coordinates a cohort-design component, authorised database access, a statistician that invokes deterministic analytical tools, and a reporting component that produces a reproducible notebook and final report (Figure~\ref{fig:use-case-cards}a). Human approval gates apply to the research question, cohort definition, analytical plan and final report. The system has read-only access to source data and does not determine patient eligibility or initiate clinical or operational actions.

\clearpage
\clearpage

Under Layer 1, the use case is assigned a base agentic risk of \(\mathbf{A}\mathbf{8}\) (Table~\ref{tab:1}) because the planner decomposes the objective and executes a goal-directed sequence across specialised components and tools. Reversibility is assigned a multiplier of 1.0 because analytical outputs can be reviewed, rejected and rerun before being incorporated into downstream decisions. Transparency is assigned a multiplier of 1.0 because the cohort specification, analytical plan, tool calls, statistical outputs and reproducible notebook provide a reconstructable execution record. Patient impact is assigned a multiplier of 1.25 because the outputs do not directly affect care or eligibility but may indirectly influence clinical-development strategy, evidence-generation priorities or treatment-landscape interpretation.

The resulting risk-adjusted score is 10.

\[\mathbf{R = A\  \times \ }\mathbf{M}_{\mathbf{reversibility}}\mathbf{\  \times \ }\mathbf{M}_{\mathbf{transparency}}\mathbf{\  \times \ }\mathbf{M}_{\mathbf{patient}}\]

\[\mathbf{R =}\mathbf{8}_{\mathbf{(agentic\ risk:\ goal\ directed\ autonomy)}}\mathbf{\  \times \ }\mathbf{1\ }_{\mathbf{(easy\ to\ reverse)}}\mathbf{\times \ }\mathbf{1}_{\mathbf{(transparent)}}\mathbf{\  \times \ }\mathbf{1.25}_{\mathbf{(indirect\ patient\ impact)}}\]

\[\mathbf{R = 10\ (High)\ }\]

The decision tree (Figure~\ref{fig:decision-tree}) would therefore route the use case to the High-risk validation pathway (Figure~\ref{fig:use-case-cards}a). The classification can be revisited if the system's scope changes, particularly if it obtains write access, triggers workflow actions, performs autonomous optimisation, or influences patient eligibility.

Under Layer 2, foundational controls include role-based access management, patient-information masking before model invocation, an approved data-source register, full production tracing, defined retention requirements for analytical artefacts and report snapshots, approved analytical tools, sandbox restrictions and cross-functional governance involving epidemiological, statistical, technical and data-governance expertise.

Under Layer 3, validation assesses cohort-design accuracy, analytical-workflow construction, statistical-method selection, tool execution, numerical correctness, and report reproducibility. Performance criteria should be evaluated against reference cohort definitions and analytical outputs where available. Testing should include structured expert adjudication, repeated-run assessment, notebook reproduction, SQL- and prompt-injection testing, sandbox-escape and blocked-import tests, malformed input or data-manifest scenarios and failures involving unavailable or inconsistent source data.

Under Layer 4, the evidence package includes the context-of-use specification, risk-classification record, data specification, cohort and variable definitions, model and component versions, analytical-tool registry, validation protocol, predefined numerical tolerances, scenario-test results, expert-adjudication records, reproducible analysis-code exports, execution logs, deviation records, monitoring plan and revalidation triggers. Revalidation can be initiated following material changes to data sources, cohort logic, models, prompts, analytical tools, orchestration, access permissions or intended use.

\subsubsection{Site selection and prioritisation}\label{site-selection-and-prioritisation}

The site-selection and prioritisation use case operates during study start-up. An orchestrator coordinates protocol parsing, generation of site-selection criteria, retrieval of authorised real-world operational, commercial and competitive-intelligence signals, site ranking and comparison of country configurations (Figure~\ref{fig:use-case-cards}b). The protocol parser, site ranker and country optimiser invoke tools and approved data sources. Final site selection and all downstream activation actions remain subject to human approval.

Under Layer 1, the use case is assigned a calibrated base agentic risk of \(\mathbf{A} = 5.5\). As it extends beyond a conventional advisory ranker, recommendations incorporate multiple contextual signals and coordinated analytical steps. Nonetheless, it does not autonomously execute downstream actions, \hl{and therefore receives a lower score than the other use cases}. Reversibility is assigned a multiplier of 1.0 because rankings can be reviewed, revised or rejected before site activation. Although ranking signals and rationale are inspectable, the heterogeneous evidence and multi-step ranking logic may limit complete contestability; transparency is therefore assigned a multiplier of 1.25., \hl{but} the heterogeneous evidence and multi-step ranking logic may limit complete contestability, transparency is assigned a multiplier of 1.25. Finally, patient impact is assigned a multiplier of 1.0 because, within the specified boundary, the system supports internal pre-activation decision-making and does not directly determine eligibility, recruitment or patient-level action.

The resulting risk-adjusted score is 6.9.

\begin{equation*}\resizebox{\linewidth}{!}{$\displaystyle \mathbf{R =}\mathbf{5.5}_{\mathbf{(agentic\ risk:\ recommend\ with\ context)}}\mathbf{\  \times \ }\mathbf{1\ }_{\mathbf{(easy\ to\ reverse)}}\mathbf{\times \ }\mathbf{1.25}_{\mathbf{(partially\ explainable)}}\mathbf{\  \times \ }\mathbf{1}_{\mathbf{(no\ plausible\ patient\ impact)}}$}\end{equation*}

\[\mathbf{R = 6.9\ (Moderate)\ }\]

This corresponds to a Moderate validation category (Figure~\ref{fig:use-case-cards}b).

Under Layer 2, foundational controls include an approved source register, documented protocol-parsing and ranking specifications, role-based access control, defined tool permissions, traceability of retrieved feasibility signals, version control for ranking logic, retention requirements for rankings and intermediate evidence, supplier-risk management and business-governance approval before deployment.

Under Layer 3, validation includes diverse protocol scenarios, comparison with historical site-selection decisions, structured clinical-operations adjudication, ranking stability under missing or conflicting data, sensitivity to criteria or weighting changes, assessment of geographic and population representation and trajectory review of the ranking and country-comparison workflow. Acceptance criteria should be specified before testing.

Under Layer 4, the evidence package includes the context-of-use specification, source and signal register, ranking specification, risk-classification rationale, validation protocol, scenario-test results, historical benchmarks, expert-adjudication records, ranking traces, deviations, monitoring metrics and revalidation triggers. Revalidation should be required after material changes to protocol parsing, data sources, ranking criteria, weighting logic, component versions, tool permissions or intended use.

\subsubsection{Enrolment forecasting and rescue recommendation}\label{enrolment-forecasting-and-rescue-recommendation}

Enrolment forecasting and rescue recommendation operates during trial conduct and links current performance to assumptions made during study start-up. An orchestrator performs task decomposition across forecasting, deviation assessment and rescue planning using adaptive evidence collection to update its assessment as operational signals become available and conditional tool use to invoke trial-performance and site-management functions according to the evolving workflow state. Across study, country, region and site levels, the system estimates enrolment trajectories, evaluates plausible causes of deviation, and proposes targeted interventions using activation dates, screening and enrolment volumes, screen-failure patterns, eligibility requirements, site capacity and competing-study activity (Figure~\ref{fig:use-case-cards}c). Recommendations may include targeted site support, additional site activation, revised recruitment approaches or escalation. The system does not execute rescue actions.

Under Layer 1, the use case is assigned base agentic risk \(\mathbf{A}\mathbf{6}\) because it generates context-aware recommendations from time-varying operational signals and inferred causes of underperformance. Reversibility, transparency, and patient-impact multipliers are each 1.25. Recommendations can be modified or withdrawn, but delayed or inappropriate intervention may have persistent operational consequences. Source signals and uncertainty estimates are reviewable, although root-cause attribution may remain contestable. Patient impact is indirect because rescue actions may affect trial access, recruitment patterns and representativeness. The risk-adjusted score is 11.7, corresponding to High (Figure~\ref{fig:use-case-cards}c).

Under Layer 2, foundational controls include an approved operational-source register, defined data-latency and completeness thresholds, role-based access, bounded tool permissions, documented forecasting and root-cause logic, retained recommendation traces, human approval before intervention, and change control for data feeds, prompts, models and orchestration.

Under Layer 3, validation assesses temporal back-testing, forecast calibration, false-alert and missed-alert rates, robustness to short-term noise and missing data, root-cause attribution, recommendation appropriateness, site-level and subgroup performance, repeated-run stability, and hand-offs between trial-review and site-management components. Scenario tests should cover activation delays, abrupt screening changes, incomplete reporting, eligibility mismatch, capacity constraints and conflicting signals.

Under Layer 4, the evidence package includes the context-of-use specification, operational-source register, forecasting and recommendation specifications, risk-classification rationale, validation protocol, back-testing and scenario results, adjudication records, recommendation traces, override and rejection records, monitoring thresholds and revalidation triggers. Revalidation should follow material changes to data feeds, protocol assumptions, models, prompts, recommendation logic, tool permissions or intended use.

\subsubsection{Trial design trade-off support}\label{trial-design-trade-off-support}

Trial-design trade-off support operates during design and planning, with hand-offs into study start-up after a design is selected. An orchestrator coordinates statistical, clinical, operational and trial-design components to compare candidate designs across statistical power, endpoints, sample size, eligibility criteria, geography, operational feasibility, timelines, cost and participant burden (Figure~\ref{fig:use-case-cards}d). The system exposes the assumptions, constraints, objective weights and sensitivity analyses underlying each option. Human approval gates apply to assumptions, weighting, scenario selection and the final design; autonomous protocol modification or commitment is prohibited.

Under Layer 1, the use case is assigned base agentic risk \(\mathbf{A}\mathbf{9}\) because it balances competing scientific, operational and participant-related objectives. Reversibility multiplier is 1.5 because design decisions become progressively difficult to unwind as protocol development advances. Transparency multiplier is 1.5 because recommendations depend on assumptions, objective weights, constraints, calculation tools and scenario-generation logic. Patient impact is 1.5 because design choices may affect eligibility, visit burden, endpoint measurement and participant access. The risk-adjusted score is 30.4, corresponding to Very High (Figure~\ref{fig:use-case-cards}d).

Under Layer 2, foundational controls include approved statistical and simulation tools, explicit objective and constraint specifications, versioned assumptions and weights, role-based permissions, prohibited autonomous protocol changes, independent clinical and statistical review, full trajectory logging and change control for models, prompts, tools and orchestration.

Under Layer 3, validation includes independent verification of calculation tools, sensitivity analyses, constraint-violation tests, dominated-option detection, repeated-run stability, subgroup and participant-burden assessment, and review of hand-offs among statistical, clinical, operational and trial-design components. Scenario testing should determine whether uncertainty and value judgements are exposed and whether the system avoids presenting a design as uniquely optimal when its ranking depends materially on subjective weighting.

Under Layer 4, the evidence package includes the context-of-use specification, design-option and constraint register, versioned assumptions and weights, risk-classification rationale, validation protocol, tool-verification records, sensitivity and scenario results, independent review records, trajectory logs, deviations, monitoring plan and revalidation triggers. Revalidation should follow material changes to endpoints, eligibility criteria, assumptions, objective weights, models, tools, orchestration or intended use.

Across the four applications, trajectory-level evaluation identified failure modes that could remain concealed by plausible final outputs. These included incorrect variable mapping or unauthorised retrieval in epidemiological analysis, silent handling of missing feasibility signals in site prioritisation, misattribution of enrolment underperformance to an incorrect operational cause, and design recommendations that depended on insufficiently exposed objective weights. These examples illustrate why output plausibility should be supplemented by reconstruction and evaluation of intermediate behaviour.

We performed sensitivity analysis to examine whether the illustrative use-case risk scores depended on the numerical values assigned to the contextual multipliers. We added a \hl{moderately} expanded multiplicative specification to increase separation between successive risk levels without altering the underlying category structure and added a compressed specification to reduce that separation (Table~\ref{tab:5}). Under the baseline model, reversibility and transparency use multipliers of 1.0, 1.25, and 1.5 across increasing risk levels, whilst patient impact uses 1.0, 1.25, 1.5, and 3.0. In the moderately expanded calibration, the intermediate and upper multipliers for reversibility and transparency were increased from 1.25 and 1.5 to 1.35 and 1.55; the corresponding patient-impact levels were increased from 1.25 and 1.5 to 1.35 and 1.55, whilst the lowest and highest patient-impact values remained 1.0 and 3.0. In the compressed calibration, the intermediate and upper multipliers for reversibility and transparency were reduced to 1.1 and 1.25; the corresponding patient-impact values were reduced to 1.1 and 1.25, with the highest patient-impact value reduced from 3.0 to 2.0 (Table~\ref{tab:5}).

\begingroup\footnotesize
\setlength{\tabcolsep}{3pt}
\begin{longtable}[]{@{}
  >{\raggedright\arraybackslash}p{(\columnwidth - 6\tabcolsep) * \real{0.2162}}
  >{\raggedright\arraybackslash}p{(\columnwidth - 6\tabcolsep) * \real{0.2861}}
  >{\raggedright\arraybackslash}p{(\columnwidth - 6\tabcolsep) * \real{0.2477}}
  >{\raggedright\arraybackslash}p{(\columnwidth - 6\tabcolsep) * \real{0.2500}}@{}}
\caption{Calibration specifications of multipliers.}\label{tab:5}\\
\toprule\noalign{}
\begin{minipage}[b]{\linewidth}\raggedright
\textbf{Calibration}
\end{minipage} & \begin{minipage}[b]{\linewidth}\raggedright
\textbf{Reversibility}
\end{minipage} & \begin{minipage}[b]{\linewidth}\raggedright
\textbf{Transparency}
\end{minipage} & \begin{minipage}[b]{\linewidth}\raggedright
\textbf{Patient impact}
\end{minipage} \\
\midrule\noalign{}
\endfirsthead
\toprule\noalign{}
\begin{minipage}[b]{\linewidth}\raggedright
\textbf{Calibration}
\end{minipage} & \begin{minipage}[b]{\linewidth}\raggedright
\textbf{Reversibility}
\end{minipage} & \begin{minipage}[b]{\linewidth}\raggedright
\textbf{Transparency}
\end{minipage} & \begin{minipage}[b]{\linewidth}\raggedright
\textbf{Patient impact}
\end{minipage} \\
\midrule\noalign{}
\endhead
\bottomrule\noalign{}
\endlastfoot
Baseline & 1.0, 1.25, 1.5 & 1.0, 1.25, 1.5 & 1.0, 1.25, 1.5, 3.0 \\
Moderately expanded & 1.0, 1.35, 1.55 & 1.0, 1.35, 1.55 & 1.0, 1.35, 1.55, 3.0 \\
Compressed & 1.0, 1.1, 1.25 & 1.0, 1.1, 1.25 & 1.0, 1.1, 1.25, 2.0 \\
\end{longtable}
\endgroup

Site selection remained the lowest-scoring and trial-design support the highest-scoring application across all specifications (Table~\ref{tab:6}). Under the compressed calibration, however, epidemiology and enrolment rescue moved from High to Moderate and trial-design support moved from Very High to High, whilst site selection remained Moderate. The intermediate ordering of epidemiology and enrolment rescue also reversed. Thus, the identity of the lowest- and highest-scoring applications was stable, whereas threshold-based category assignment was more sensitive to calibration. This supports organisational calibration of both multipliers and category thresholds.

\begingroup\footnotesize
\setlength{\tabcolsep}{3pt}
\begin{longtable}[]{@{}
  >{\raggedright\arraybackslash}p{(\columnwidth - 6\tabcolsep) * \real{0.2971}}
  >{\raggedright\arraybackslash}p{(\columnwidth - 6\tabcolsep) * \real{0.2407}}
  >{\raggedright\arraybackslash}p{(\columnwidth - 6\tabcolsep) * \real{0.2436}}
  >{\raggedright\arraybackslash}p{(\columnwidth - 6\tabcolsep) * \real{0.2185}}@{}}
\caption{Sensitivity of use-case risk scores to alternative calibration specifications}\label{tab:6}\\
\toprule\noalign{}
\begin{minipage}[b]{\linewidth}\raggedright
\textbf{Use case}
\end{minipage} & \begin{minipage}[b]{\linewidth}\raggedright
\textbf{\hl{Baseline calibration}}
\end{minipage} & \begin{minipage}[b]{\linewidth}\raggedright
\textbf{Moderately expanded calibration}
\end{minipage} & \begin{minipage}[b]{\linewidth}\raggedright
\textbf{Compressed calibration}
\end{minipage} \\
\midrule\noalign{}
\endfirsthead
\toprule\noalign{}
\begin{minipage}[b]{\linewidth}\raggedright
\textbf{Use case}
\end{minipage} & \begin{minipage}[b]{\linewidth}\raggedright
\textbf{\hl{Baseline calibration}}
\end{minipage} & \begin{minipage}[b]{\linewidth}\raggedright
\textbf{Moderately expanded calibration}
\end{minipage} & \begin{minipage}[b]{\linewidth}\raggedright
\textbf{Compressed calibration}
\end{minipage} \\
\midrule\noalign{}
\endhead
\bottomrule\noalign{}
\endlastfoot
Epidemiology and treatment-landscape analysis & 8 \hl{\(\times\)} 1.0 \hl{\(\times\)} 1.0 \hl{\(\times\)} 1.25 =

\textbf{10.0 - High} & 8 \hl{\(\times\)} 1.0 \hl{\(\times\)} 1.0 \hl{\(\times\)} 1.35 =

\textbf{10.8 - High} & 8 \hl{\(\times\)} 1.0 \hl{\(\times\)} 1.0 \hl{\(\times\)} 1.1 =

\textbf{8.8 - Moderate} \\
Site selection and prioritisation & 5.5 \hl{\(\times\)} 1.0 \hl{\(\times\)} 1.25 \hl{\(\times\)} \hl{1.0} =

\textbf{6.9 - Moderate} & 5.5 \hl{\(\times\)} 1.0 \hl{\(\times\)} 1.35 \hl{\(\times\)} \hl{1.0} =

\textbf{7.4 - Moderate} & 5.5 \hl{\(\times\)} 1.0 \hl{\(\times\)} 1.1 \hl{\(\times\)} \hl{1.0} =

\textbf{6.1 - Moderate} \\
Enrolment forecasting and rescue recommendation & 6 \hl{\(\times\)} 1.25 \hl{\(\times\)} 1.25 \hl{\(\times\)} 1.25 =

\textbf{11.7 - High} & 6 \hl{\(\times\)} 1.35 \hl{\(\times\)} 1.35 \hl{\(\times\)} 1.35 =

\textbf{14.8 - High} & 6 \hl{\(\times\)} 1.1 \hl{\(\times\)} 1.1 \hl{\(\times\)} 1.1 =

\textbf{8.0 - Moderate} \\
Trial-design trade-off support & 9 \hl{\(\times\)} 1.5 \hl{\(\times\)} 1.5 \hl{\(\times\)} 1.5 =

\textbf{30.4 - Very High} & 9 \hl{\(\times\)} 1.55 \hl{\(\times\)} 1.55 \hl{\(\times\)} 1.55 =

\textbf{33.5 - Very High} & 9 \hl{\(\times\)} 1.25 \hl{\(\times\)} 1.25 \hl{\(\times\)} 1.25 =

\textbf{17.6 - High} \\
\end{longtable}
\endgroup

\section{Discussion}\label{discussion}

VITAL translates regulatory and use-case evidence into an operational method for trajectory-based, risk-proportionate validation of agentic AI in pharmaceutical workflows and clinical trial operations (Figures~\ref{fig:vital-framework} and \ref{fig:standards-crosswalk}). Our framework reframes validation as an assessment of bounded, observable system behaviour. Deterministic validation asks whether a system returns the expected result for a defined input; agentic validation asks whether a system pursues an intended objective through acceptable steps, information sources, tools, hand-offs and human-control points.

Adjacent approaches address different assurance questions and can therefore contribute different forms of evidence to an agentic validation decision (Supplementary Note 4; Supplementary Table~\ref{tab:s6}). Task-specific medical-agent benchmarks provide performance evidence under defined evaluation conditions\textsuperscript{35}, healthcare-oriented testing and evaluation approaches such as CHAI provide structured evidence concerning agent behaviour and workflow performance\textsuperscript{15,16}, VECTR addresses reliability in drug-development applications\textsuperscript{36}, and established CSV and CSA practices remain relevant to controlled software functions and lifecycle assurance\textsuperscript{12,29}. For pharmaceutical validation, however, the issue is the applicability of such evidence to the intended use. Evidence generated through an adjacent approach should be carried forward only where the evaluated system configuration, data environment, tool permissions, workflow boundary and acceptance criteria are sufficiently aligned with the system being validated. VITAL therefore provides a context for combining heterogeneous assurance evidence without assuming equivalence between the frameworks or processes from which that evidence originated. This preserves the relationship between each validation claim, the evidence supporting it and the system state under which that evidence was obtained.

This interpretation reinforces three implications for agentic AI validation. First, the unit of validation is the trajectory as well as the output: intermediate steps can appear plausible whilst producing unsafe or misleading system-level behaviour\hl{, so the trajectory should be captured to enable reconstruction of system behaviour, which} supports both validation and ongoing oversight. Second, assurance is socio-technical \hl{in that} workflow integration, ownership, escalation, user review and governance are part of validation itself. Third, evidence should be risk-proportionate where higher autonomy, lower reversibility, weaker transparency or greater patient consequence justify more in-depth testing, stronger oversight and more intensive monitoring. A risk-based approach mitigates the limitations of a uniform validation approach and aligns validation effort with system impact.

VITAL is implementation-agnostic but not architecture-neutral. Higher-risk agentic systems cannot be validated unless the deployment environment can constrain autonomy, record trajectories, expose failure modes, manage dependencies, and re-test material changes. Enterprise systems should separate orchestration, model serving, retrieval, tool access, guardrails, telemetry, evaluation and operations into distinct architectural planes. The orchestration plane defines agent roles, task decomposition, hand-offs, stopping rules and escalation logic. The model-serving plane defines model versions, inference endpoints and release state. The retrieval plane defines approved data sources, indexing, embedding, reranking, citation and provenance controls. The tool-access plane defines authorised APIs, parameter limits, write permissions, sandboxing and blocked actions. The guardrail plane constrains inputs, retrieval, dialogue, tool execution and outputs. The telemetry and evaluation plane records traces, metrics, user interactions, benchmark results and regression tests. The operations plane governs deployment, access, cost, latency, monitoring, incident response and decommissioning.

Security controls should be enforced independently of model compliance and across the complete execution path. Higher-risk systems should combine least-privilege access, network-egress restrictions, credential isolation, resource and time limits, tamper-\hl{evident} telemetry and the ability to suspend or terminate a run. Evaluations that weaken production safeguards should be treated as a distinct context of use and should test whether persistent goal pursuit can combine individually permissible actions to circumvent system instructions\textsuperscript{37--39}.

We make several recommendations based on our implementation of the VITAL framework. For example, telemetry should be designed before production use. At minimum, each run should capture the initiating user or workflow, task objective, system instructions, retrieved sources, intermediate plans, tool calls, parameters, tool outputs, errors, retries, guardrail activations, model and agent versions, final outputs, human review actions and downstream actions. For operational monitoring, telemetry should also capture latency, token use, compute utilisation, cost, overrides, edits, abandoned outputs, rejected recommendations and escalations. We also suggest that evaluation should be continuous. Generic benchmarks can support model selection, but pharmaceutical workflows require deployment-specific scenario sets reflecting common cases, boundary cases, missing data, conflicting evidence, rare but consequential failures and foreseeable misuse. Higher-risk agents should undergo scenario testing, adversarial testing, repeated-run assessment, retrieval-quality assessment, tool-call validation, human-review testing and regression testing after material changes. Automated evaluators, cross-agent checking and teacher-student review can improve scale and resilience, but they should supplement rather than replace subject-matter expert adjudication where outputs influence clinical development decisions.

Responsibility should follow architecture. Model providers may own model runtime, release documentation and security updates. Infrastructure providers may own deployment tooling, lifecycle support, resource controls and operational resilience. Implementation teams may own orchestration, retrieval configuration, tool integration, telemetry and workflow embedding. Sponsors remain accountable for intended use, data suitability, risk acceptance, user training, process integration and operational oversight. Quality and governance functions should approve validation strategy, residual risk, deviation handling and revalidation rules.

Recent incident disclosures illustrate why validation should encompass evaluation conditions and infrastructure as well as production workflows. During an internal cyber-capability evaluation, OpenAI reported that its models exploited a zero-day vulnerability in an internally hosted proxy, obtaining access to the broader internet and compromising the infrastructure of a third party (i.e., Hugging Face)\textsuperscript{38,39}. The case supports treating sandboxing and model-level instructions as fallible controls whose effectiveness should be demonstrated across complete trajectories. Cases like this further support the need for adequate validation and tracking systems that allow organisations to detect, explain and mitigate future events. Evaluation environments, runtime dependencies, network routes and credentials should fall within the validated system boundary.

VITAL is relevant to regulatory, quality and enterprise oversight because it converts agentic execution into auditable evidence that can support failure attribution, impact assessment and proportionate corrective action across clinical development and adjacent pharmaceutical domains including RWE, manufacturing and financial-control processes. The regulatory and standards mapping identified a gap in operational guidance for agentic AI; approaches such as VITAL could help inform how regulators and standards bodies address this gap in three areas. First, guidance should define when an AI system becomes agentic for validation purposes. Relevant features include task decomposition, tool access, workflow orchestration, inter-agent hand-offs, memory, write capability, adaptive behaviour and the ability to influence or execute downstream actions. Second, guidance should specify the minimum evidence required for higher-risk workflows, including intended-use boundaries, permitted and prohibited actions and reconstructable execution trajectories. Changes to models, tools and retrieval corpora should be evaluated against their effects on system behaviour. Third, accountability should be allocated across sponsors, technology providers, and platform vendors for data provenance, system configuration, tool permissions, telemetry, incident response, change control, and residual-risk acceptance. These considerations do not necessarily require a separate regulatory pathway for agents, but indicate where existing AI, software, GxP and clinical-trial quality principles require greater operational specificity.

Agentic AI is likely to enter pharmaceutical research and development through progressively more complex decision support, workflow orchestration and bounded automation. VITAL provides a lifecycle-oriented, trajectory-based and risk-proportionate approach for defining the minimum evidence required to validate these systems within a specified context of use. However, we acknowledge several limitations. The numerical parameters for risk adjustment are illustrative and not empirically derived, and sensitivity analysis indicates that classification of \hl{several} applications can vary under alternative calibrations (Tables~\ref{tab:5} and \ref{tab:6}\hl{)}. Another limitation concerns the structure of the risk model itself. The multiplicative formulation treats autonomy, reversibility, transparency and patient impact as separable dimensions whose combined influence can be represented through fixed scalar adjustments. In practice, however, these dimensions may interact. For example, limited transparency may have greater consequence when an action is difficult to reverse than when the same output remains advisory, whilst patient impact may depend on the position of an agentic action within a longer decision chain rather than on the action in isolation. The present formulation should therefore be interpreted as an operational decision aid. Future empirical work should examine whether interaction terms, non-linear effects or use-case-specific weighting improve classification stability. Federated and cross-organisational agentic workflows where control and accountability may be distributed across multiple entities were also outside the present scope.

The current applications demonstrate how VITAL can be operationalised but do not yet establish the performance of the framework as a validation method. We therefore plan a 12--18 month prospective mixed-methods evaluation of VITAL across 3--5 agentic AI systems in at least two organisations (AstraZeneca and IQVIA). Primary endpoints may include inter-rater reliability of risk-tier assignment amongst independent validators, time to validation decision relative to baseline CSV practice, and the proportion of prespecified behavioural failure modes detected during Layer 3 validation. Secondary endpoints \hl{may} consider validator-reported usability, completeness against Layer 4 evidence requirements, and downstream quality or regulatory audit findings, with interim analysis planned at six months. We think prospective evaluation should additionally examine the responsiveness of VITAL to controlled changes in system configuration. A validation framework should distinguish changes that alter the assurance problem from changes that do not. This could be tested by applying VITAL before and after predefined modifications to model version, retrieval corpus, tool permissions, orchestration logic or human-control placement with assessors blinded to whether each modification was expected to alter risk. Appropriate upward or downward changes in validation requirements would provide evidence of sensitivity to consequential system change, whereas frequent reclassification after immaterial modifications would indicate poor specificity and unnecessary validation burden. 

\section{Methods}\label{methods}

We developed an agentic AI validation framework using a multi-component approach integrating landscape analysis from the literature, mapping and analysis of regulations and standards, subject-matter expert interviews, and iterative framework refinement.

\subsection{Landscape analysis}\label{landscape-analysis-1}

A targeted landscape analysis was undertaken to identify validation challenges, candidate functional dimensions and emerging evaluation approaches relevant to agentic AI in pharmaceutical development settings. Searches were conducted across peer-reviewed literature, regulatory and standards documents, grey literature, and relevant industry sources.

Search activity focused on three areas: operational bottlenecks in drug development; scientific and technical work relevant to agentic or large language model-enabled approaches in healthcare and drug development; and validation of agentic systems.

Sources were analysed for recurring validation challenges, cross-cutting themes, functional validation dimensions, candidate evaluation approaches and implications for validation framework design in GxP-adjacent settings (Supplementary Note 1).

\subsection{Standards and regulatory frameworks mapping}\label{standards-and-regulatory-frameworks-mapping}

A structured mapping exercise was undertaken to identify the validation expectations relevant to agentic AI and to determine where the agentic validation framework would require coverage. The mapping included seven regulatory groupings spanning GxP/CSV/Part 11\textsuperscript{17}; FDA CSA\textsuperscript{12} and GAMP 5\textsuperscript{29}; International Medical Device Regulators Forum (IMDRF) Good Machine Learning Practice (GMLP)\textsuperscript{18}; ISO/IEC 42001\textsuperscript{19} and ISO/IEC 23894\textsuperscript{20}; the EU AI Act\textsuperscript{21} and ICH E6(R3)\textsuperscript{22}; FDA/EMA Guiding principles of good AI Practice for drug development\textsuperscript{23}; and EU GMP Annex 22\textsuperscript{24}.

For each grouping, regulatory intent was evaluated and analysis was performed to identify the degree to which a standard includes guidance (coverage) for aspects most relevant to agentic systems, including intended use, risk classification, change control, human oversight, lifecycle monitoring, supplier management and transparency or auditability. Coverage was classified using predefined criteria as full, partial, implied or gap. Full coverage required an explicit operational requirement applicable to the domain; partial coverage indicated a relevant principle without implementation detail; implied coverage indicated indirect or narrow applicability; and gap indicated no substantive requirement. Findings were used to define the evaluation domains in the framework.

\subsection{Subject matter expert interviews}\label{subject-matter-expert-interviews}

A subject-matter expert interview study was undertaken to capture technical, operational and governance perspectives on the validation of agentic AI in pharmaceutical workflows (Supplementary Note 2). Eighteen subject-matter experts were recruited using purposive expert sampling on the basis of relevant domain expertise and experience with the design, deployment or oversight of agentic AI in operationally sensitive settings. Participants represented technical development, data science, product and platform development, clinical development, clinical operations, governance and adjacent translational roles. Eighteen participants were approached by email in January 2026. All participants accepted the interview invitation. Interviews were conducted between January \hl{and} March 2026 by A.L. and R.C.M. who had relevant experience in clinical research, AI and data science. The interviewers had no prior relationship with participants. Before participation, interviewees were informed of the study purpose, the interviewers\textquotesingle{} roles and how their responses would be used in framework development. Interviews were conducted virtually via the Microsoft Teams platform and lasted approximately 30 to 45 minutes. Transcripts were returned to participants. No non-participants were present. Interviews were recorded and transcribed using Microsoft Teams. Field notes were made during or immediately after interviews. No repeat interviews were conducted. Sample composition was designed to provide coverage across technical, operational, clinical and governance perspectives relevant to framework development; recruitment was not governed by a thematic-saturation stopping rule.

A semi-structured interview guide was developed iteratively to support consistent coverage of the principal topics whilst allowing sufficient flexibility for participants to elaborate on their role-specific \hl{perspective}\textsuperscript{40,41} (Supplementary Table~\ref{tab:s4}). The interview guide was pilot tested with four individuals (A.L., R.C.M., R.S., S.M.) and refined before full data collection. The guide combined exploratory questions with targeted probes and scenario-based follow-up questions to elicit examples of expected value, workflow pain points, potential failure modes, evidence required for sign-off, perceived risk, and leadership or governance considerations. Interviews were transcribed and analysed using a hybrid deductive-inductive framework \hl{analysis}\textsuperscript{42}. This approach was selected because the study had a set of pre-specified analytic interests derived from the broader research objectives, whilst also aiming to identify additional concerns emerging directly from the interview data. An initial analytic framework was derived from study aims and included domains relating to use-case selection, risk, autonomy, reversibility, transparency, patient consequence, benchmarking, governance, traceability and monitoring.

Transcript coding was undertaken by \hl{A.L.,} R.C.M. and S.M. Transcripts were read in full and were subsequently coded by assigning segments of text to one or more analytic categories within the framework. The coding framework was refined iteratively as additional concepts emerged from the data. Coded material was then charted in a matrix by participant and analytic domain to support systematic comparison across interviews and stakeholder groups. To preserve participant and organisational confidentiality including obligations arising from contractual non-disclosure agreements, interview findings are reported in synthesised form and not \hl{as} attributable quotations. Higher-order themes were developed through examination of recurrent patterns, areas of convergence and points of divergence relevant to validation framework design and operationalisation. This approach is consistent with uses of framework analysis in multidisciplinary applied health research and with hybrid deductive-inductive qualitative analysis more \hl{broadly}\textsuperscript{43,44}. Reporting of this interview component was checked against the Consolidated Criteria for Reporting Qualitative Research (COREQ) with applicable items reported in the Supplementary Information\textsuperscript{45}.

\subsection{Validation framework development and refinement}\label{validation-framework-development-and-refinement}

Framework development was conducted iteratively, integrating evidence from the literature, the standards and regulatory mapping and analysis, and expert interviews. Structured multi-stakeholder workshops and working sessions were used to review emerging concepts, align terminology, define framework components, and examine the relevance of candidate evaluation domains. These sessions brought together pharmaceutical clinical development, clinical operations, quality and governance, data science, product, platform and infrastructure perspectives to ensure that the framework addresses both regulatory expectations and implementation constraints.

\subsection{Structured application to four pharmaceutical use cases}\label{structured-application-to-four-pharmaceutical-use-cases}

Four use cases were specified using a common template covering intended use, lifecycle stage, component roles, data and tool access, human-control points, prohibited actions and plausible consequences. Base agentic risk and contextual multipliers were assigned using the risk ladder and risk-adjustment model. The resulting category was then used to derive foundational controls, behavioural tests, evidence outputs, monitoring requirements and revalidation triggers.

\textbf{\ul{Technical scope of the use-case applications}}

The four use cases comprised two existing agentic systems developed within AstraZeneca and two illustrative applications to demonstrate broader framework applicability. The epidemiology and treatment-landscape analysis (Figure~\ref{fig:use-case-cards}a) and site selection and prioritisation (Figure~\ref{fig:use-case-cards}b) use cases were based on actual agentic systems. Both incorporated large language model components to support task interpretation, planning and orchestration alongside deterministic analytical tools and authorised data services. To protect commercially sensitive information, these systems are described through non-confidential abstractions covering intended use, agent roles, orchestration, authorised data and tool access, human-control points, prohibited actions and traceability requirements. Source code and proprietary datasets are not reported.

\textbf{\ul{Data availability}}

The sources underlying the landscape and standards analyses are identified in the article and supplementary information. Interview transcripts and proprietary technical materials are not publicly available because their disclosure is restricted by participant confidentiality, organisational confidentiality and contractual non-disclosure obligations. The article therefore reports only aggregate findings and non-confidential abstractions of the relevant systems.

\section{Acknowledgements}\label{acknowledgements}

We gratefully acknowledge the contributions of the subject matter experts who participated in interviews and provided critical insights that informed the development and refinement of the validation framework. We thank Jonathan Green, Christina Mack, Cheng Qian, Katie Bonner, Etai Jacob, Steve Kiddle, Andrzej Patyra, Nils Svangard, Andrew Osborne, Asad Rahman, Nina Mian, Phil Teare, and Marta Tolos for sharing their expertise across clinical, technical and governance domains. This work was supported by AstraZeneca through a collaborative project with IQVIA \& NVIDIA. No separate grant funding was received for this work.

\section{Author Contributions}\label{author-contributions}

Conceptualization: S.B., R.S.

Methodology: S.B., \hl{P.M.F.}, A.H.A., A.T., S.M., A.L., R.C.M., R.S.

Investigation: S.B., \hl{P.M.F.}, A.H.A., A.T., S.M., A.L., R.C.M., R.S.

Visualization: \hl{P.M.F.}, A.H.A., S.M., A.L., R.C.M.

Project administration: A.L., R.C.M.

Data curation: \hl{P.M.F.}, A.H.A., S.M., A.L., R.C.M.

Validation: S.B., \hl{P.M.F.}, A.H.A., A.T., X.Y., R.M., S.M., A.L., R.C.M., R.S.

Formal analysis: \hl{P.M.F.}, A.H.A., S.M., A.L., R.C.M.

Resources: S.B., R.S.

Funding Acquisition: S.B., R.S.

Writing - original draft: A.L., R.C.M.

Writing - review and editing: S.B., \hl{P.M.F.}, A.H.A., A.T., A.G., N.J., X.Y., R.M., S.M., A.L., R.C.M., P.S., R.S.

\section{Competing interests}\label{competing-interests}

\hl{P.M.F.} and N.J. are employees of TekSystems and receive a salary from TekSystems, but their work is contracted out to AstraZeneca. The authors declare no other competing interests.

\section{References}\label{references}

1. FDA. Framework for FDA's real-world evidence program. https://www.fda.gov/media/120060/download (2018).

2. FDA. Considerations for the Use of Real-World Data and Real-World Evidence To Support Regulatory Decision-Making for Drug and Biological Products. https://www.fda.gov/regulatory-information/search-fda-guidance-documents/considerations-use-real-world-data-and-real-world-evidence-support-regulatory-decision-making-drug (2023).

3. Qiu, J. \emph{et al.} LLM-based agentic systems in medicine and healthcare. \emph{Nat. Mach. Intell.} \textbf{6}, 1418--1420 (2024).

4. Zou, J. \& Topol, E. J. The rise of agentic AI teammates in medicine. \emph{The Lancet} \textbf{405}, 457 (2025).

5. FDA. FDA Expands Artificial Intelligence Capabilities with Agentic AI Deployment. https://www.fda.gov/news-events/press-announcements/fda-expands-artificial-intelligence-capabilities-agentic-ai-deployment (2025).

6. NIST. Artificial Intelligence Risk Management Framework (AI RMF 1.0). (2023).

7. FDA. Adaptive Design Clinical Trials for Drugs and Biologics Guidance for Industry. https://www.fda.gov/regulatory-information/search-fda-guidance-documents/adaptive-design-clinical-trials-drugs-and-biologics-guidance-industry (2019).

8. FDA. Digital Health Technologies for Remote Data Acquisition in Clinical Investigations. https://www.fda.gov/regulatory-information/search-fda-guidance-documents/digital-health-technologies-remote-data-acquisition-clinical-investigations (2023).

9. FDA. Artificial Intelligence for Drug Development. https://www.fda.gov/about-fda/center-drug-evaluation-and-research-cder/artificial-intelligence-drug-development (2026).

10. GOV UK. Complying with consumer law when using AI agents. https://www.gov.uk/government/publications/complying-with-consumer-law-when-using-ai-agents/complying-with-consumer-law-when-using-ai-agents (2026).

11. Collaco, B. G. \emph{et al.} The role of agentic artificial intelligence in healthcare: a scoping review. \emph{Npj Digit. Med.} \textbf{9}, 345 (2026).

12. FDA. Computer Software Assurance for Production and Quality Management System Software. https://www.fda.gov/regulatory-information/search-fda-guidance-documents/computer-software-assurance-production-and-quality-management-system-software (2026).

13. Bandi, A., Kongari, B., Naguru, R., Pasnoor, S. \& Vilipala, S. V. The rise of agentic ai: A review of definitions, frameworks, architectures, applications, evaluation metrics, and challenges. \emph{Future Internet} \textbf{17}, 404 (2025).

14. Gridach, M., Nanavati, J., Abidine, K. Z. E., Mendes, L. \& Mack, C. Agentic ai for scientific discovery: A survey of progress, challenges, and future directions. \emph{ArXiv Prepr. ArXiv250308979} (2025).

15. Coalition for Health AI. Best Practice Guide - Agentic AI (v1.0). https://www.chai.org/workgroup/use-case/agentic (2026).

16. Coalition for Health AI. Testing \& Evaluation (T\&E) Framework - Agentic AI (v1.0). https://www.chai.org/workgroup/use-case/agentic (2026).

17. FDA. Guidance for industry: Part 11, electronic records; electronic signatures - scope and application. https://www.fda.gov/media/75414/download (2003).

18. International Medical Device Regulators Forum. Good machine learning practice for medical device development: Guiding principles. (2025).

19. ISO/IEC. ISO/IEC 42001:2023. https://www.iso.org/standard/42001 (2023).

20. ISO/IEC. ISO/IEC 23894:2023. https://www.iso.org/standard/77304.html (2023).

21. EU AI Act. High-level summary of the AI Act. https://artificialintelligenceact.eu/high-level-summary/ (2024).

22. ICH. ICH Harmonised Guideline For Good Clinical Practice E6(R3). https://database.ich.org/sites/default/files/ICH\_E6\%28R3\%29\_Step4\_FinalGuideline\_2025\_0106.pdf (2025).

23. FDA/EMA. Guiding Principles of Good AI Practice in Drug Development. https://www.fda.gov/about-fda/artificial-intelligence-drug-development/guiding-principles-good-ai-practice-drug-development (2026).

24. European Commission. Annex 22: Artificial Intelligence. https://health.ec.europa.eu/document/download/5f38a92d-bb8e-4264-8898-ea076e926db6\_en?filename=mp\_vol4\_chap4\_annex22\_consultation\_guideline\_en.pdf (2025).

25. Jia, F., Wu, T., Qin, X. \& Squicciarini, A. The task shield: Enforcing task alignment to defend against indirect prompt injection in llm agents. in \emph{Proceedings of the 63rd Annual Meeting of the Association for Computational Linguistics (Volume 1: Long Papers)} 29680--29697 (2025).

26. Xiang, Z. \emph{et al.} Guardagent: safeguard LLM agents via knowledge-enabled reasoning. in \emph{ICML 2025 workshop on computer use agents} (2025).

27. Lu, J. \emph{et al.} Toolsandbox: A stateful, conversational, interactive evaluation benchmark for llm tool use capabilities. in \emph{Findings of the Association for Computational Linguistics: NAACL 2025} 1160--1183 (2025).

28. Zhu, K. \emph{et al.} Multiagentbench: Evaluating the collaboration and competition of llm agents. in \emph{Proceedings of the 63rd Annual Meeting of the Association for Computational Linguistics (Volume 1: Long Papers)} 8580--8622 (2025).

29. ISPE. ISPE GAMP® 5: A Risk-Based Approach to Compliant GxP Computerized Systems (Second Edition). https://guidance-docs.ispe.org/doi/book/10.1002/9781946964571 (2022).

30. Hulstaert, L., Twick, I., Sarsour, K. \& Verstraete, H. Enhancing site selection strategies in clinical trial recruitment using real-world data modeling. \emph{Plos One} \textbf{19}, e0300109 (2024).

31. Ni, Y. A multi-agent AI platform for clinical trial operations: Specialized agents to accelerate feasibility assessment while ensuring data accuracy. (2026).

32. Jin, L. \emph{et al.} Large Language Models in Randomized Controlled Trials Design: Observational Study. \emph{J. Med. Internet Res.} \textbf{27}, e67469 (2025).

33. Safin, D. \& Balta, D. Autonomy and Agency in Agentic AI: Architectural Tactics for Regulated Contexts. \emph{ArXiv Prepr. ArXiv260512105} (2026).

34. U.S. Securities and Exchange Commission. Management's Report on Internal Control Over Financial Reporting and Certification of Disclosure in Exchange Act Periodic Reports. https://www.sec.gov/rules-regulations/2003/03/managements-report-internal-control-over-financial-reporting-certification-disclosure-exchange-act (2003).

35. Jiang, Y. \emph{et al.} MedAgentBench: a virtual EHR environment to benchmark medical LLM agents. \emph{Nejm Ai} \textbf{2}, AIdbp2500144 (2025).

36. Lagarde, W. VECTR: Towards a Reliability Framework for Agentic AI in Drug Development. \emph{Preprint} https://zenodo.org/records/18122849 (2026).

37. OpenAI. Safety and alignment in an era of long-horizon models. https://openai.com/index/safety-alignment-long-horizon-models/ (2026).

38. OpenAI. OpenAI and Hugging Face partner to address security incident during model evaluation. https://openai.com/index/hugging-face-model-evaluation-security-incident/ (2026).

39. Hugging Face. Security incident disclosure - July 2026. https://huggingface.co/blog/security-incident-july-2026 (2026).

40. DeJonckheere, M. \& Vaughn, L. M. Semistructured interviewing in primary care research: a balance of relationship and rigour. \emph{Fam. Med. Community Health} \textbf{7}, e000057 (2019).

41. Kallio, H., Pietilä, A.-M., Johnson, M. \& Kangasniemi, M. Systematic methodological review: developing a framework for a qualitative semi-structured interview guide. \emph{J. Adv. Nurs.} \textbf{72}, 2954--2965 (2016).

42. Proudfoot, K. Inductive/deductive hybrid thematic analysis in mixed methods research. \emph{J. Mix. Methods Res.} \textbf{17}, 308--326 (2023).

43. Fereday, J. \& Muir-Cochrane, E. Demonstrating rigor using thematic analysis: A hybrid approach of inductive and deductive coding and theme development. \emph{Int. J. Qual. Methods} \textbf{5}, 80--92 (2006).

44. Gale, N. K., Heath, G., Cameron, E., Rashid, S. \& Redwood, S. Using the framework method for the analysis of qualitative data in multi-disciplinary health research. \emph{BMC Med. Res. Methodol.} \textbf{13}, 117 (2013).

45. Tong, A., Sainsbury, P. \& Craig, J. Consolidated criteria for reporting qualitative research (COREQ): a 32-item checklist for interviews and focus groups. \emph{Int. J. Qual. Health Care} \textbf{19}, 349--357 (2007).
\clearpage
\begingroup
\begin{center}
{\large\bfseries VITAL: A framework for validating agentic AI in pharmaceutical drug development and clinical trial operations}
\end{center}
\setcounter{table}{0}\renewcommand{\tablename}{Supplementary Table}
\section*{Supplementary Information}
\addcontentsline{toc}{section}{Supplementary Information}
\emph{\textbf{\ul{Supplementary Note 1. Targeted landscape analysis informing development of the VITAL framework}}}

\emph{\ul{Purpose.}} A targeted landscape analysis was undertaken to identify concepts requiring operational treatment in a validation framework for agentic AI in clinical development and clinical operations. The analysis was intended to support framework development by characterising recurrent validation problems, candidate behavioural domains and emerging evaluation methods. The comparatively limited literature directly addressing agentic AI in clinical development and clinical operations also required inclusion of evidence from adjacent healthcare, pharmaceutical, technical and regulated settings.

\emph{\ul{Search strategy and source selection}}. An a priori keyword framework was developed across three concept groups. Agentic-system terms included ``agentic AI'', ``agentic system*'', ``AI agent*'', ``autonomous agent*'', ``LLM agent*'', ``multi-agent'', ``agentic workflow'' and ``agentic automation''. Validation and assurance terms included ``validation'', ``evaluation'', ``verification'', ``assurance'', ``benchmarking'', ``reliability'', ``robustness'', ``safety'', ``trustworthiness'', ``audit*'', ``traceability'', ``reproducibility'', ``governance'', ``human oversight'' and ``human-in-the-loop''. Clinical-development and clinical-operations terms included ``clinical trial*'', ``clinical development'', ``clinical operations'' and ``trial operation*''.

Searches were conducted in PubMed and Google Scholar, the latter for broader identification of papers, preprints and relevant material not indexed in PubMed. Regulatory, standards and industry sources relevant to validation and assurance were identified through searches of FDA, ISO/IEC, EU AI Act, GxP, computer-system validation and computer-software-assurance materials.

Titles were retained when they indicated agentic behaviour together with a validation or evaluation dimension, or direct relevance to assurance in regulated settings. Abstracts were then assessed to determine whether the source addressed agentic behaviour, contributed validation-relevant methods, metrics, failure modes or governance controls, and was applicable to healthcare, clinical research or a transferable regulated setting. Full-text review was prioritised for sources meeting these criteria.

\emph{\ul{Evidence base.}} The final evidence base comprised 113 sources selected for full-text review (Supplementary References). It included peer-reviewed journal and conference publications, regulatory and standards documents, preprints and grey literature. Sources addressing healthcare, pharmaceutical development, regulated software and adjacent assurance settings were retained where the validation problem was considered transferable to agentic workflows in clinical development or clinical operations.

The composition of the evidence base is reported in Supplementary Table~\ref{tab:s1}.

\begingroup\footnotesize
\setlength{\tabcolsep}{3pt}
\begin{longtable}[]{@{}
  >{\raggedright\arraybackslash}p{(\columnwidth - 4\tabcolsep) * \real{0.5667}}
  >{\raggedright\arraybackslash}p{(\columnwidth - 4\tabcolsep) * \real{0.2167}}
  >{\raggedright\arraybackslash}p{(\columnwidth - 4\tabcolsep) * \real{0.2166}}@{}}
\caption{Composition of the landscape-analysis evidence base. Grey literature included industry reports, editorials and white papers.}\label{tab:s1}\\
\toprule\noalign{}
\begin{minipage}[b]{\linewidth}\raggedright
Source category
\end{minipage} & \begin{minipage}[b]{\linewidth}\raggedright
Sources, n
\end{minipage} & \begin{minipage}[b]{\linewidth}\raggedright
Proportion
\end{minipage} \\
\midrule\noalign{}
\endfirsthead
\toprule\noalign{}
\begin{minipage}[b]{\linewidth}\raggedright
Source category
\end{minipage} & \begin{minipage}[b]{\linewidth}\raggedright
Sources, n
\end{minipage} & \begin{minipage}[b]{\linewidth}\raggedright
Proportion
\end{minipage} \\
\midrule\noalign{}
\endhead
\bottomrule\noalign{}
\endlastfoot
\emph{Peer-reviewed journal and conference publications} & 72 & 63.7\% \\
\emph{Regulatory guidance and standards} & 15 & 13.3\% \\
\emph{Preprints} & 14 & 12.4\% \\
\emph{Grey literature} & 12 & 10.6\% \\
\emph{Total} & \textbf{113} & \textbf{100\%} \\
\end{longtable}
\endgroup

\emph{\ul{Analytical procedure.}} The included sources were examined around four analytic categories:

\begin{itemize}
\item
  validation challenge classes
\item
  cross-cutting implications for assurance
\item
  agent capabilities requiring direct behavioural evaluation
\item
  candidate methods for generating validation evidence
\end{itemize}

Individual sources could address multiple challenges or capabilities. Broader themes were then developed to explain how those problems altered the unit, scope or evidentiary requirements of validation.

\emph{\ul{Framework-relevant findings}}

\begin{enumerate}
\def\labelenumi{\arabic{enumi}.}
\item
  Validation challenges and cross-cutting implications
\end{enumerate}

Ten validation challenge classes were identified and are summarised in Supplementary Table~\ref{tab:s2}. Their implications were consolidated into eight cross-cutting themes.

\begingroup\footnotesize
\setlength{\tabcolsep}{3pt}
\begin{longtable}[]{@{}
  >{\raggedright\arraybackslash}p{(\columnwidth - 4\tabcolsep) * \real{0.5214}}
  >{\raggedright\arraybackslash}p{(\columnwidth - 4\tabcolsep) * \real{0.2659}}
  >{\raggedright\arraybackslash}p{(\columnwidth - 4\tabcolsep) * \real{0.2127}}@{}}
\caption{Validation challenge classes identified in the landscape analysis. Categories were non-mutually exclusive.}\label{tab:s2}\\
\toprule\noalign{}
\begin{minipage}[b]{\linewidth}\raggedright
Validation challenge class
\end{minipage} & \begin{minipage}[b]{\linewidth}\raggedright
Sources, n
\end{minipage} & \begin{minipage}[b]{\linewidth}\raggedright
Proportion
\end{minipage} \\
\midrule\noalign{}
\endfirsthead
\toprule\noalign{}
\begin{minipage}[b]{\linewidth}\raggedright
Validation challenge class
\end{minipage} & \begin{minipage}[b]{\linewidth}\raggedright
Sources, n
\end{minipage} & \begin{minipage}[b]{\linewidth}\raggedright
Proportion
\end{minipage} \\
\midrule\noalign{}
\endhead
\bottomrule\noalign{}
\endlastfoot
\emph{Model drift and continuous learning} & 98 & 87\% \\
\emph{Integration with regulated workflows} & 94 & 83\% \\
\emph{Documentation and evidence artefacts} & 92 & 81\% \\
\emph{Data access controls and confidentiality} & 92 & 81\% \\
\emph{Scalability, workflow fit and adoption} & 89 & 79\% \\
\emph{Hallucination and factuality} & 57 & 50\% \\
\emph{Non-determinism and reproducibility} & 42 & 37\% \\
\emph{Multi-agent decomposition} & 40 & 35\% \\
\emph{Security, prompt injection and tool misuse} & 14 & 12\% \\
\emph{Autonomy boundaries and approval gates} & 5 & 4\% \\
\end{longtable}
\endgroup

\begin{enumerate}
\def\labelenumi{\arabic{enumi}.}
\setcounter{enumi}{1}
\item
  System-level performance cannot be inferred from component performance
\end{enumerate}

The evidence indicated that individually capable components may interact in ways that reduce overall system reliability. This was relevant to multi-agent architectures, in which errors, assumptions or incomplete information may be transferred across hand-offs. Validation needed to assess collective behaviour, orchestration and information flow rather than relying on component-level tests.

\begin{enumerate}
\def\labelenumi{\arabic{enumi}.}
\setcounter{enumi}{2}
\item
  Reproducibility requires behavioural rather than exact-output criteria
\end{enumerate}

Agentic systems may follow different acceptable paths when responding to equivalent inputs. Exact replication of wording or intermediate actions was not always an appropriate validation criterion. The analysis supported evaluation of semantic consistency, process integrity, constraint adherence and the stability of consequential outcomes across repeated runs.

\begin{enumerate}
\def\labelenumi{\arabic{enumi}.}
\setcounter{enumi}{3}
\item
  Plausible outputs may conceal invalid execution
\end{enumerate}

Linguistic fluency and apparent coherence were insufficient indicators of validity. A credible final output could arise from unsupported evidence, inappropriate retrieval, incorrect tool use or unacceptable intermediate reasoning.

\begin{enumerate}
\def\labelenumi{\arabic{enumi}.}
\setcounter{enumi}{4}
\item
  Agentic validation requires additional evidence
\end{enumerate}

Conventional validation records did not consistently capture the information required to reconstruct agent behaviour. Relevant outputs included retrieved-source records, tool calls, intermediate plans, hand-off records, prompt and orchestration versions, guardrail activations, human interventions and execution outcomes.

\begin{enumerate}
\def\labelenumi{\arabic{enumi}.}
\setcounter{enumi}{5}
\item
  Validation must accommodate lifecycle change
\end{enumerate}

Agent behaviour may change following updates to foundation models, prompts, retrieval sources, tools, orchestration logic or external dependencies. The evidence supported lifecycle monitoring, controlled change assessment and risk-based revalidation.

\begin{enumerate}
\def\labelenumi{\arabic{enumi}.}
\setcounter{enumi}{6}
\item
  Exhaustive testing is generally infeasible
\end{enumerate}

The combination of variable inputs, tool states, execution paths and environmental conditions creates a large and potentially open-ended test space. This supported the use of risk-based scenario selection covering representative cases, boundary conditions, foreseeable misuse and rare but consequential failures.

\begin{enumerate}
\def\labelenumi{\arabic{enumi}.}
\setcounter{enumi}{7}
\item
  Errors can propagate across otherwise valid steps
\end{enumerate}

An early error may be accepted and transformed by subsequent components, producing a technically coherent but contextually invalid result. This supported evaluation of decision-critical assumptions, hand-off integrity and cumulative error propagation across the complete workflow.

\begin{enumerate}
\def\labelenumi{\arabic{enumi}.}
\setcounter{enumi}{8}
\item
  Human oversight is a control requiring validation
\end{enumerate}

Human review may be weakened by automation bias, workload, incomplete information or poorly positioned approval gates. Human-in-the-loop arrangements were therefore treated as workflow controls whose timing, effectiveness and failure modes required direct evaluation.

\emph{\ul{Derivation of behavioural validation domains}}

Five recurring capability areas were identified as requiring direct behavioural assessment. Their frequency within the evidence base is reported in Supplementary Table~\ref{tab:s3}.

\begingroup\footnotesize
\setlength{\tabcolsep}{3pt}
\begin{longtable}[]{@{}
  >{\raggedright\arraybackslash}p{(\columnwidth - 6\tabcolsep) * \real{0.3182}}
  >{\raggedright\arraybackslash}p{(\columnwidth - 6\tabcolsep) * \real{0.1666}}
  >{\raggedright\arraybackslash}p{(\columnwidth - 6\tabcolsep) * \real{0.1667}}
  >{\raggedright\arraybackslash}p{(\columnwidth - 6\tabcolsep) * \real{0.3484}}@{}}
\caption{Functional validation dimensions identified in the landscape analysis. The dimensions were candidate organising constructs and were subsequently refined through the other framework-development activities.}\label{tab:s3}\\
\toprule\noalign{}
\begin{minipage}[b]{\linewidth}\raggedright
Functional dimension
\end{minipage} & \begin{minipage}[b]{\linewidth}\raggedright
Sources, n
\end{minipage} & \begin{minipage}[b]{\linewidth}\raggedright
Proportion
\end{minipage} & \begin{minipage}[b]{\linewidth}\raggedright
Evaluation focus
\end{minipage} \\
\midrule\noalign{}
\endfirsthead
\toprule\noalign{}
\begin{minipage}[b]{\linewidth}\raggedright
Functional dimension
\end{minipage} & \begin{minipage}[b]{\linewidth}\raggedright
Sources, n
\end{minipage} & \begin{minipage}[b]{\linewidth}\raggedright
Proportion
\end{minipage} & \begin{minipage}[b]{\linewidth}\raggedright
Evaluation focus
\end{minipage} \\
\midrule\noalign{}
\endhead
\bottomrule\noalign{}
\endlastfoot
\emph{Reasoning and planning} & 96 & 85\% & Completeness, ordering and appropriateness of planned actions \\
\emph{Grounding and retrieval} & 93 & 82\% & Relevance, provenance, synthesis and attribution of retrieved evidence \\
\emph{Human interaction and governance} & 90 & 80\% & Oversight, interpretability, interruption, escalation and safe use \\
\emph{Tool use and execution fidelity} & 52 & 46\% & Validity, authorisation and intended alignment of external actions \\
\emph{Multi-agent coordination} & 40 & 35\% & Hand-off integrity, conflict resolution and collective behaviour \\
\end{longtable}
\endgroup

\begin{enumerate}
\def\labelenumi{\arabic{enumi}.}
\item
  Reasoning and planning
\end{enumerate}

This domain concerned whether an agent decomposed an objective into a complete, coherent and appropriately ordered sequence of actions. Relevant failure modes included omitted steps, inappropriate sequencing, unsupported assumptions, ineffective stopping rules and failure to escalate uncertainty.

\begin{enumerate}
\def\labelenumi{\arabic{enumi}.}
\setcounter{enumi}{1}
\item
  Tool use and execution
\end{enumerate}

This domain concerned whether interactions with databases, analytical software, application programming interfaces and other external systems were valid, authorised and aligned with the intended task. Attention was given to the boundary at which probabilistic agent decisions were converted into deterministic system actions, because an apparently reasonable intention could produce an invalid or harmful external action through an incorrect tool call or parameter.

\begin{enumerate}
\def\labelenumi{\arabic{enumi}.}
\setcounter{enumi}{2}
\item
  Grounding and retrieval
\end{enumerate}

This domain concerned whether the system retrieved, synthesised and attributed information from relevant, authorised and sufficiently reliable sources. Relevant failure modes included unsupported synthesis, inaccurate citation, incomplete retrieval and inappropriate handling of conflicting evidence.

\begin{enumerate}
\def\labelenumi{\arabic{enumi}.}
\setcounter{enumi}{3}
\item
  Multi-agent coordination
\end{enumerate}

This domain concerned the integrity of communication and collective behaviour across specialised components. Evaluation considerations included information loss, corruption or reinterpretation during hand-offs, unresolved disagreement, inconsistent assumptions and system-level behaviour not apparent from individual component performance.

\begin{enumerate}
\def\labelenumi{\arabic{enumi}.}
\setcounter{enumi}{4}
\item
  Human-agent interaction
\end{enumerate}

This domain concerned whether users could understand, supervise, interrupt and act safely upon agent behaviour. Relevant considerations included interpretability, escalation, override effectiveness, reviewer workload and the capture of human decisions within the execution record.

These five capability areas provided the initial organising structure for the behavioural validation domains in VITAL. Their final definitions were refined through regulatory and standards mapping, subject-matter expert interviews and multi-stakeholder review.

\emph{\ul{Emerging evaluation methods}}

Four methodological approaches were identified as potentially applicable to agentic-system validation.

\begin{enumerate}
\def\labelenumi{\arabic{enumi}.}
\item
  Verification-aware planning
\end{enumerate}

Verification checks may be embedded within planning and execution. Deterministic checks are suited to structured calculations, schemas and tool outputs, whereas semantic checks may be required for factual consistency, completeness and adherence to contextual constraints.

\begin{enumerate}
\def\labelenumi{\arabic{enumi}.}
\setcounter{enumi}{1}
\item
  Simulation-based evaluation
\end{enumerate}

Controlled environments can be used to assess retrieval, tool execution, state changes, permission boundaries and error recovery without exposing production systems. Simulation also permits repeatable evaluation of unsafe or low-frequency scenarios that may be difficult to test operationally.

\begin{enumerate}
\def\labelenumi{\arabic{enumi}.}
\setcounter{enumi}{2}
\item
  Trajectory evaluation
\end{enumerate}

Trajectory-based methods examine intermediate plans, retrieved evidence, tool calls, hand-offs and control activations alongside final outputs. This enables identification of unsafe shortcuts, unsupported assumptions, inefficient loops and unacceptable execution pathways that would not be detected through output review alone.

\begin{enumerate}
\def\labelenumi{\arabic{enumi}.}
\setcounter{enumi}{3}
\item
  Automated evaluation
\end{enumerate}

Model-based adjudication and perturbation testing may increase the scale of regression and scenario testing. Their use requires calibration against independent reference evidence and periodic subject-matter expert review. Automated evaluation was considered complementary to, rather than a replacement for, domain adjudication.

\emph{\ul{Contribution to framework development}}

The landscape analysis informed VITAL in six ways. First, non-determinism and open-ended workflow states supported risk-proportionate testing. Second, multi-step execution, cumulative error propagation and the probabilistic-to-deterministic action boundary supported treating the execution trajectory alongside the final output as a unit of validation. Third, the five capability areas informed the behavioural validation domains. Fourth, recurrent concerns relating to data governance, tool permissions, autonomy boundaries, security and human oversight informed the foundational control requirements. Fifth, the absence of standardised agentic evidence outputs informed the framework's traceability, logging and documentation requirements. Sixth, model drift, changing dependencies and evolving workflow conditions informed the monitoring, change-control and revalidation provisions.

\emph{\textbf{\ul{\hfill\break
}}}

\emph{\textbf{\ul{Supplementary Note 2. Subject-matter expert interview topic-domain guide}}}

A semi-structured topic-domain guide was used to support data collection across subject-matter expert (SME) interviews. The guide was developed to ensure coverage of pre-specified domains relevant to agentic AI validation in pharmaceutical research and development, whilst allowing interviewers to pursue role-specific elaboration and emergent concerns. The domains below summarise the areas explored during the interviews.

\begingroup\footnotesize
\setlength{\tabcolsep}{3pt}
\begin{longtable}[]{@{}
  >{\raggedright\arraybackslash}p{(\columnwidth - 4\tabcolsep) * \real{0.2651}}
  >{\raggedright\arraybackslash}p{(\columnwidth - 4\tabcolsep) * \real{0.3827}}
  >{\raggedright\arraybackslash}p{(\columnwidth - 4\tabcolsep) * \real{0.3522}}@{}}
\caption{Subject-matter expert interview topic domains.}\label{tab:s4}\\
\toprule\noalign{}
\begin{minipage}[b]{\linewidth}\raggedright
\textbf{Topic domain}
\end{minipage} & \begin{minipage}[b]{\linewidth}\raggedright
\textbf{Scope of inquiry}
\end{minipage} & \begin{minipage}[b]{\linewidth}\raggedright
\textbf{Relevance to framework development}
\end{minipage} \\
\midrule\noalign{}
\endfirsthead
\toprule\noalign{}
\begin{minipage}[b]{\linewidth}\raggedright
\textbf{Topic domain}
\end{minipage} & \begin{minipage}[b]{\linewidth}\raggedright
\textbf{Scope of inquiry}
\end{minipage} & \begin{minipage}[b]{\linewidth}\raggedright
\textbf{Relevance to framework development}
\end{minipage} \\
\midrule\noalign{}
\endhead
\bottomrule\noalign{}
\endlastfoot
Orientation to agentic AI in clinical development and clinical operations & Expert perspectives on perceived opportunities, uncertainties and concerns associated with agentic AI in clinical development and clinical operations. & Established the contextual assumptions, perceived value propositions and risk concerns that shaped subsequent exploration of validation requirements. \\
Operational pain points and candidate use cases & Workflow bottlenecks and candidate applications in which agentic AI could plausibly support complex, repeatable or multi-step tasks, including feasibility assessment, site selection, recruitment, monitoring, trial master file processes, query management and safety triage. & Informed the framework's use-case-specific orientation and reinforced the need to define validation relative to an intended workflow context. \\
Distinction between agentic AI and conventional analytics & Characteristics distinguishing agentic systems from dashboards, predictive models or conventional analytics, including planning, retrieval, tool use, task coordination, recommendations and bounded action. & Supported the framework's focus on agentic behaviour as the object of validation, including reasoning, planning, tool invocation, workflow boundaries and human-control points. \\
Use-case suitability and generalisability & Features that make a use case suitable for framework application, including generalisability across organisations or therapeutic areas, feasibility of test design, availability of metrics and ability to construct an evidence package. & Contributed to the selection and framing of representative validation scenarios and clarified the need for framework components that can accommodate heterogeneous pharmaceutical workflows. \\
Failure modes and delegation boundaries & Potential failures associated with agent involvement, including silent errors, inappropriate autonomy, cumulative error propagation, unsafe escalation, regulatory consequences, patient-safety implications and activities unsuitable for delegation. & Informed the framework's treatment of intrinsic risk, autonomy limits, prohibited actions, escalation criteria and risk-proportionate validation depth. \\
Confidence and evidence requirements & Evidence required by functional, technical or governance stakeholders before accepting an agentic system in a production or operationally sensitive workflow, including test scenarios, acceptance criteria, auditability, reproducibility and reviewability. & Informed the evidence-layer structure of the framework, including validation protocols, evidence packages, audit trails and traceability requirements. \\
Functional implementation constraints & Conditions required for deployment in clinical development or clinical operations, including data availability, data quality, workflow maturity, system integration, user review processes and observability. & Supported the framework's foundational control requirements, particularly data governance, operational readiness, telemetry, integration controls and human oversight. \\
Risk tiering and regulated-workflow relevance & Factors influencing perceived risk level, including autonomy, write access, affected systems, reversibility, proximity to regulated records, patient consequence and GxP relevance. & Informed the risk-tiering logic and the principle that validation effort should scale with autonomy, reversibility, transparency and potential patient or operational consequence. \\
Continuity with existing validation practices & Aspects of computer system validation, computer software assurance, good machine learning practice and related quality approaches that remain applicable to agentic AI, as well as limitations when applied to non-deterministic, tool-using systems. & Supported the framework's positioning as a translation of existing principles into agentic AI-specific validation domains. \\
Platform architecture and technical feasibility & Technical factors influencing feasibility, including orchestration patterns, agent frameworks, tool-use architecture, latency, security, data locality, environment separation and platform constraints. & Informed the framework's treatment of agent design controls, tool permissions, technical boundaries, third-party dependencies and architectural prerequisites. \\
Observability, telemetry and trace reconstruction & Mechanisms for observing and reconstructing agent behaviour, including logging of prompts, inputs, retrieved sources, tool calls, parameters, intermediate plans, retries, error states, outputs, model or system versions and user-workflow linkage. & Contributed to traceability and auditability requirements, including the need to reconstruct trajectories. \\
Integration patterns and responsibility boundaries & Interactions between agents and customer systems, internal data lakes, document repositories, application programming interfaces, connectors and data pipelines, together with responsibility allocation across vendors, platform providers and deploying organisations. & Informed third-party and supplier management requirements, responsibility allocation, access management, fallback arrangements and risk segmentation. \\
Validation design for agentic behaviour & Methods for demonstrating that an agentic system performs acceptably within a defined context of use, including scenario-based testing, simulation, edge-case testing, adversarial inputs, historical comparison, expert adjudication, experiment tracking and comparison of system variants. & Informed behavioural validation domains, including reasoning and planning, tool use and execution, retrieval, coordination, human-agent interaction and revalidation triggers. \\
Leadership stance and organisational readiness & Leadership expectations for safe adoption of agentic AI, including communication of benefits and risks, human accountability, transparency, escalation, skills, training, process maturity, tooling, infrastructure and organisational culture. & Supported the framework's socio-technical framing by identifying organisational readiness, leadership accountability and user capability as relevant to validation and deployment. \\
Governance roles and lifecycle ownership & Allocation of responsibility for framework ownership, validation approval, ongoing monitoring, change control and oversight across clinical operations, quality assurance, data science, machine learning, information technology and product functions. & Informed the governance operating model, lifecycle accountability and the need for distributed ownership across business, technical and quality functions. \\
\end{longtable}
\endgroup

The guide was applied flexibly according to participant expertise. Functional SMEs were primarily engaged on operational pain points, workflow feasibility, evidence expectations, risk tiering, regulated-process relevance and applicability of current validation practices. Technical SMEs were additionally engaged on architecture, tool-use patterns, telemetry, integration mechanisms, security constraints, environment separation and platform capabilities required to support validation in regulated or operationally sensitive contexts. This role-specific adaptation preserved comparability at the level of core domains whilst enabling depth in areas where participants held specialist knowledge.

\emph{\textbf{\ul{\hfill\break
}}}

\emph{\textbf{\ul{Supplementary Note 3. Minimum record for reconstruction of a VITAL validation decision}}}

The evidence generated through VITAL may be distributed across risk assessments, control specifications, behavioural test results, expert adjudications, deviation records, monitoring plans and technical artefacts. Layer 4 defines the categories of evidence that should be retained, but the existence of a complete evidence package does not by itself make the resulting validation judgement readily reconstructable. A later reviewer may need to determine not only what evidence was available, but which findings influenced the decision, how conflicting or incomplete evidence was treated, and which assumptions or controls were necessary for the resulting disposition to remain valid.

Supplementary Table~\ref{tab:s5} therefore defines a minimum validation-decision record intended to sit above the Layer 4 evidence package. The record provides an account of the inferential step between the assembled evidence and the final governance disposition.

\begingroup\footnotesize
\setlength{\tabcolsep}{3pt}
\begin{longtable}[]{@{}
  >{\raggedright\arraybackslash}p{(\columnwidth - 4\tabcolsep) * \real{0.2541}}
  >{\raggedright\arraybackslash}p{(\columnwidth - 4\tabcolsep) * \real{0.3588}}
  >{\raggedright\arraybackslash}p{(\columnwidth - 4\tabcolsep) * \real{0.3870}}@{}}
\caption{Minimum validation-decision record for VITAL applications.}\label{tab:s5}\\
\toprule\noalign{}
\begin{minipage}[b]{\linewidth}\raggedright
\textbf{Record element}
\end{minipage} & \begin{minipage}[b]{\linewidth}\raggedright
\textbf{Purpose}
\end{minipage} & \begin{minipage}[b]{\linewidth}\raggedright
\textbf{Minimum information to retain}
\end{minipage} \\
\midrule\noalign{}
\endfirsthead
\toprule\noalign{}
\begin{minipage}[b]{\linewidth}\raggedright
\textbf{Record element}
\end{minipage} & \begin{minipage}[b]{\linewidth}\raggedright
\textbf{Purpose}
\end{minipage} & \begin{minipage}[b]{\linewidth}\raggedright
\textbf{Minimum information to retain}
\end{minipage} \\
\midrule\noalign{}
\endhead
\bottomrule\noalign{}
\endlastfoot
\textbf{Validation disposition} & Defines precisely what has been authorised or not authorised & The resulting disposition and its scope, including any restriction on users, workflows, decisions, data, tools or downstream actions \\
\textbf{Decision claims} & Identifies the propositions on which the disposition depends & The small set of claims that must be true for the disposition to remain justified with references to the evidence supporting each claim \\
\textbf{Decision-critical evidence} & Distinguishes evidence that influenced the judgement from evidence that was merely available & The principal test results, expert judgements, control evidence or other findings that materially affected the disposition referenced to their sources \\
\textbf{Adverse or conflicting evidence} & Prevents unfavourable findings from being obscured by an aggregate conclusion & Material failed tests, contradictory findings, unresolved deviations or expert disagreement, together with their effect on the final judgement \\
\textbf{Uncertainty and assumptions} & Makes explicit where the decision relies on incomplete knowledge or judgement & Material uncertainties, assumptions or areas in which available evidence was insufficient to provide a definitive conclusion and the consequence of that uncertainty for use \\
\textbf{Decision conditionality} & Identifies controls or operating conditions on which acceptability depends & Controls, restrictions, human interventions or environmental conditions whose continued effectiveness is necessary for the disposition to remain valid \\
\textbf{Residual-risk rationale} & Records why remaining risk was considered acceptable, conditionally acceptable or unacceptable & The material residual risks remaining after controls and the rationale for accepting, constraining or rejecting them \\
\textbf{Validity envelope} & Defines the state of the system to which the decision applies & The system configuration and contextual assumptions within which the evidence and resulting disposition remain applicable with references to configuration-controlled records where available \\
\textbf{Decision invalidation criteria} & Distinguishes routine monitoring from circumstances requiring reconsideration of the original decision & Specific changes, observations or failures that would render one or more decision-critical claims unsupported and therefore require the validation disposition to be reopened \\
\textbf{Decision provenance} & Preserves accountability for the final judgement & Decision date, approving function or authority, substantive dissent where present and references to the approved decision record \\
\end{longtable}
\endgroup

The minimum validation-decision record should be updated when the validation disposition changes rather than whenever underlying evidence is modified. This distinction preserves the relationship between the decision and the evidence state on which it was based. Where subsequent information invalidates a decision-critical claim, the existing record should be retained as the historical basis for the earlier disposition and a new or revised disposition generated following reassessment.

\emph{\textbf{\ul{Supplementary Note 4. Use of evidence generated through adjacent approaches}}}

Validation of an agentic system may draw on evidence generated through more than one activity. Existing testing, reliability and computer-system assurance approaches can produce evidence relevant to particular aspects of a VITAL assessment, but their applicability depends on the conditions under which that evidence was generated. We compare adjacent approaches according to the evidence they can contribute and the conditions governing its use (Supplementary Table~\ref{tab:s6}).

For VITAL, evidence generated outside the immediate validation exercise should be assessed against four considerations. Scope concordance concerns whether the original evaluation addressed a sufficiently similar intended use, workflow and consequence. Configuration concordance concerns whether the relevant model, retrieval sources, tools, permissions and orchestration correspond to the system configuration being validated. Temporal validity concerns whether subsequent changes to the system or its dependencies have altered the conditions represented by the evidence. Claim traceability requires the evidence to be linked to the specific validation proposition it is being used to support. These considerations provide a basis for determining whether existing evidence remains applicable without assuming equivalence between the approaches through which it was generated.

\begingroup\footnotesize
\setlength{\tabcolsep}{3pt}
\begin{longtable}[]{@{}
  >{\raggedright\arraybackslash}p{(\columnwidth - 8\tabcolsep) * \real{0.1785}}
  >{\raggedright\arraybackslash}p{(\columnwidth - 8\tabcolsep) * \real{0.2367}}
  >{\raggedright\arraybackslash}p{(\columnwidth - 8\tabcolsep) * \real{0.1986}}
  >{\raggedright\arraybackslash}p{(\columnwidth - 8\tabcolsep) * \real{0.1875}}
  >{\raggedright\arraybackslash}p{(\columnwidth - 8\tabcolsep) * \real{0.1986}}@{}}
\caption{Evidentiary contribution of assurance approaches to a VITAL validation assessment.}\label{tab:s6}\\
\toprule\noalign{}
\begin{minipage}[b]{\linewidth}\raggedright
\textbf{Approach}
\end{minipage} & \begin{minipage}[b]{\linewidth}\raggedright
\textbf{Characteristic assurance evidence}
\end{minipage} & \begin{minipage}[b]{\linewidth}\raggedright
\textbf{Potential evidentiary use within VITAL}
\end{minipage} & \begin{minipage}[b]{\linewidth}\raggedright
\textbf{Conditions affecting applicability}
\end{minipage} & \begin{minipage}[b]{\linewidth}\raggedright
\textbf{Boundary of interpretation}
\end{minipage} \\
\midrule\noalign{}
\endfirsthead
\toprule\noalign{}
\begin{minipage}[b]{\linewidth}\raggedright
\textbf{Approach}
\end{minipage} & \begin{minipage}[b]{\linewidth}\raggedright
\textbf{Characteristic assurance evidence}
\end{minipage} & \begin{minipage}[b]{\linewidth}\raggedright
\textbf{Potential evidentiary use within VITAL}
\end{minipage} & \begin{minipage}[b]{\linewidth}\raggedright
\textbf{Conditions affecting applicability}
\end{minipage} & \begin{minipage}[b]{\linewidth}\raggedright
\textbf{Boundary of interpretation}
\end{minipage} \\
\midrule\noalign{}
\endhead
\bottomrule\noalign{}
\endlastfoot
\textbf{VITAL} & Context-of-use and risk assessment, documented control state, behavioural validation findings and lifecycle evidence & Provides the integrated evidence basis for determining acceptability within the specified pharmaceutical use & Evidence should correspond to the defined system boundary, validated configuration and intended use & A VITAL assessment does not replace specialised technical testing or establish compliance with external frameworks not explicitly assessed \\
\textbf{CHAI Agentic AI Testing and Evaluation Framework} & Testing and evaluation results generated against healthcare-oriented agent assessment criteria & May provide supporting evidence for relevant behavioural or human-agent interaction claims & Applicability depends on concordance of the evaluated task, system configuration, workflow conditions and acceptance criteria with the VITAL use case & Successful performance under the original evaluation conditions does not independently establish the VITAL risk classification or adequacy of the complete pharmaceutical validation evidence package \\
\textbf{VECTR} & Reliability-oriented evidence concerning agentic AI in drug-development applications & May provide supporting evidence for reliability, evidentiary integrity or traceability claims relevant to behavioural validation and documentation & The assessed agent behaviour, evidence sources and system state should remain representative of the configuration under VITAL assessment & Reliability evidence for a particular behaviour or workflow does not independently establish that all context-specific controls or validation requirements have been satisfied \\
\textbf{CSV/CSA/GAMP 5-based assurance} & Requirements, verification and testing records, traceability, controlled-change documentation and associated software-assurance evidence & May provide evidence for foundational system controls, deterministic components, data integrity, configuration management and lifecycle change & Evidence should remain applicable to the components and configuration incorporated within the agentic system and should reflect their current validated state & Component or software assurance does not independently demonstrate acceptable non-deterministic agent behaviour across alternative execution trajectories \\
\end{longtable}
\endgroup

\textbf{\ul{\hfill\break
}}

\textbf{\ul{List of abbreviations}}

Abbreviations used in the manuscript and supplementary information.

\begingroup\footnotesize
\setlength{\tabcolsep}{3pt}
\begin{longtable}[]{@{}
  >{\raggedright\arraybackslash}p{(\columnwidth - 2\tabcolsep) * \real{0.1768}}
  >{\raggedright\arraybackslash}p{(\columnwidth - 2\tabcolsep) * \real{0.8232}}@{}}
\toprule\noalign{}
\begin{minipage}[b]{\linewidth}\raggedright
\textbf{Abbreviation}
\end{minipage} & \begin{minipage}[b]{\linewidth}\raggedright
\textbf{Definition}
\end{minipage} \\
\midrule\noalign{}
\endhead
\bottomrule\noalign{}
\endlastfoot
\textbf{API} & Application programming interface \\
\textbf{CHAI} & Coalition for Health AI \\
\textbf{CSA} & Computer Software Assurance \\
\textbf{CSV} & Computer System Validation \\
\textbf{EMA} & European Medicines Agency \\
\textbf{EU} & European Union \\
\textbf{FDA} & US Food and Drug Administration \\
\textbf{GAMP 5} & Good Automated Manufacturing Practice 5 \\
\textbf{GMLP} & Good Machine Learning Practice \\
\textbf{GMP} & Good Manufacturing Practice \\
\textbf{GxP} & Good x Practice; collective term for regulated good-practice requirements \\
\textbf{ICH} & International Council for Harmonisation of Technical Requirements for Pharmaceuticals for Human Use \\
\textbf{IMDRF} & International Medical Device Regulators Forum \\
\textbf{ISO/IEC} & International Organization for Standardization/International Electrotechnical Commission \\
\textbf{RAG} & Retrieval-augmented generation \\
\textbf{RWD} & Real-world data \\
\textbf{RWE} & Real-world evidence \\
\textbf{SME} & Subject-matter expert \\
\textbf{VITAL} & Validation by Intended use, Traceability, Autonomy and Lifecycle evidence \\
\end{longtable}
\endgroup

\emph{\textbf{\ul{\hfill\break
}}}

\emph{\textbf{\ul{Supplementary References}}}

1. Mahmood F. A benchmarking crisis in biomedical machine learning. Nat Med. 2025 Apr;31(4):1060--1060.

2. Xu G, Li X, Chen Y, Duan Y, Wu S, Yu A, et al. A Comprehensive Survey of Agentic AI in Healthcare. Authorea Preprints. 2025;

3. Liu F, Niu Y, Zhang Q, Wang K, Dong Z, Wong IN, et al. A foundational architecture for AI agents in healthcare. Cell Reports Medicine. 2025 Oct;6(10):102374.

4. Laurent A. A GAMP 5 Guide to AI/ML Validation in GxP Environments. IntuitionLabs.ai. 2026;

5. Lobentanzer S, Feng S, Bruderer N, Maier A, The BioChatter Consortium, Díaz AG, et al. A platform for the biomedical application of large language models. Nat Biotechnol. 2025 Feb;43(2):166--9.

6. Serapio-García G, Safdari M, Crepy C, Sun L, Fitz S, Romero P, et al. A psychometric framework for evaluating and shaping personality traits in large language models. Nat Mach Intell. 2025 Dec 18;7(12):1954--68.

7. Banerjie S, Zhu Y, Freeman I, Machado JV, Ahmed A, Sarker A, et al. Agentic AI in Healthcare: A Comprehensive Survey of Foundations, Taxonomy, and Applications. Authorea Preprints.

8. Rao Suura S. Agentic AI Systems in Organ Health Management: Early Detection of Rejection in Transplant Patients. J Neonatal Surg. 2025 Feb 24;14(4S):490--500.

9. Raheem T, Hossain G. Agentic AI Systems: Opportunities, Challenges, and Trustworthiness. In: 2025 IEEE International Conference on Electro Information Technology (eIT). Valparaiso, IN, USA: IEEE; 2025. p. 618--24. Available from: https://ieeexplore.ieee.org/document/11103638/

10. Huang K, editor. Agentic AI: Theories and Practices. Cham: Springer Nature Switzerland; 2025. (Progress in IS). Available from: https://link.springer.com/10.1007/978-3-031-90026-6

11. Rengasamy B. Agentic Artificial Intelligence for Prior Authorization Workflows in Healthcare. Journal of Computer Science and Technology Studies. 2025;

12. Shehab MA. Agentic-AI Healthcare: Multilingual, Privacy-First Framework with MCP Agents. arXiv; 2025. Available from: http://arxiv.org/abs/2510.02325

13. Hughes L, Dwivedi YK, Malik T, Shawosh M, Albashrawi MA, Jeon I, et al. AI Agents and Agentic Systems: A Multi-Expert Analysis. Journal of Computer Information Systems. 2025 Jul 4;65(4):489--517.

14. Grammenos G, Vrahatis AG, Lazaros K, Exarchos TP, Vlamos P, Krokidis MG. AI agents in Alzheimer's disease management: challenges and future directions. Front Aging Neurosci. 2026 Jan 5;17:1735892.

15. Gorenshtein A, Omar M, Glicksberg BS, Nadkarni GN, Klang E. AI Agents in Clinical Medicine: A Systematic Review. medRxiv\,: the preprint server for health sciences. United States; 2025. p. 2025.08.22.25334232.

16. Sapkota R, Roumeliotis KI, Karkee M. AI Agents vs. Agentic AI: A Conceptual taxonomy, applications and challenges. Information Fusion. 2026 Feb;126:103599.

17. Qiu J, Wu J, Langlotz C, Huang B, Lei Z, Wu H, et al., editors. AI for Clinical Applications: First International Workshops, Agentic AI 2025, CREATE 2025, and Clinical MLLMs 2025, Held in Conjunction with MICCAI 2025, Daejeon, South Korea, September 23 and 27, 2025, Proceedings. Cham: Springer Nature Switzerland; 2026. (Lecture Notes in Computer Science; vol. 16147). Available from: https://link.springer.com/10.1007/978-3-032-06004-4

18. Shaktah LA, Carrero ZI, Hewitt KJ, Gustav M, Cecchini M, Foersch S, et al. Application of artificial intelligence and digital tools in cancer pathology. The Lancet Digital Health. 2025 Oct;7(10):100933.

19. Huynh AL, Roy TJ, Jackson KN, Lee AG, Liaw W, Hossain MM. Applications of artificial intelligence-based conversational agents in healthcare: A systematic umbrella review. International Journal of Medical Informatics. 2026 Mar;207:106204.

20. Liawrungrueang W. Artificial Intelligence (AI) Agents Versus Agentic AI: What's the Effect in Spine Surgery? Neurospine. 2025 Jun 30;22(2):473--7.

21. Ménard T, Bramstedt KA. Artificial intelligence agent in clinical trial operations: a fictional (for now) case study. AI Ethics. 2025 Oct;5(5):4627--33.

22. Panjwani S, Wei H, Mason J. Artificial intelligence and machine learning‐assisted digital applications for biopharmaceutical manufacturing. Biotechnology Progress. 2025 Nov 10;e70089.

23. Duffourc MN, Verhees FG, Gilbert S. Artificial intelligence characters are dangerous without legal guardrails. Nat Hum Behav. 2025 Dec 5; Available from: https://www.nature.com/articles/s41562-025-02375-3

24. Eda J, Mohaidat T, Niloy SI, Niu Z, Khan MRK, Khalil K. Artificial intelligence in healthcare: A comprehensive survey on AI-driven diagnosis and patient monitoring. Applied Soft Computing. 2025 Dec;114429.

25. Bignami E, Darhour LJ, Buhre W, Cecconi M, Bellini V. Artificial intelligence in healthcare: Tailoring education to meet EU AI-Act standards. Health Policy and Technology. 2025 Dec;14(6):101078.

26. Verlingue L, Boyer C, Olgiati L, Brutti Mairesse C, Morel D, Blay JY. Artificial intelligence in oncology: ensuring safe and effective integration of language models in clinical practice. The Lancet Regional Health - Europe. 2024 Nov;46:101064.

27. Durlach P, Hertogh K, Adams W, Feeney E, Ball C. At the Frontier: Gauging Health Care's Readiness for Agentic AI Innovation. NEJM AI Sponsored. 2025

28. Garcia P. Aversion to external feedback suffices to ensure agent alignment. Sci Rep. 2024 Sep 10;14(1):21147.

29. Janssen E, McLagan R, Habeck J, Chung SY, McArthur EC, Anderson P. Barriers to breakthroughs: A scoping review of generative AI in healthcare simulation. Clinical Simulation in Nursing. 2025 Oct;107:101791.

30. Dutta Roy TP. Bioethics Artificial Intelligence Advisory (BAIA): An Agentic Artificial Intelligence (AI) Framework for Bioethical Clinical Decision Support. Cureus. 2025 Mar 12; Available from: https://www.cureus.com/articles/347765-bioethics-artificial-intelligence-advisory-baia-an-agentic-artificial-intelligence-ai-framework-for-bioethical-clinical-decision-support

31. Esmaeilzadeh P. Challenges and strategies for wide-scale artificial intelligence (AI) deployment in healthcare practices: A perspective for healthcare organizations. Artificial Intelligence in Medicine. 2024 May;151:102861.

32. Warrier A. Collaborative Agentic AI for Personalized Treatment Protocol Optimization: Autonomous Clinical Decision Networks. In: International Conference on Recent Trends in Computer Science and Information Technology. ScienceTech Xplore; 2025. Available from: https://ijetcsit.org/index.php/ijetcsit/article/view/442

33. FDA. Considerations for the Use of Artificial Intelligence To Support Regulatory Decision-Making for Drug and Biological Products. 2025;

34. Moritz M, Topol E, Rajpurkar P. Coordinated AI agents for advancing healthcare. Nat Biomed Eng. 2025 Apr 1;9(4):432--8.

35. Gibney E. DeepMind unveils `spectacular' general-purpose science AI. Nature. 2025;

36. Lee C, Kumar S, Vogt KA, Munshi M, Tallapudi P, Vogt A, et al. Democratizing cost-effective, agentic artificial intelligence to multilingual medical summarization through knowledge distillation. Sci Rep. 2025 Jul 29;15(1):27619.

37. Moriuchi E. Deploying artificial intelligence for healthcare: An empirical study on the patients' usage of health care service and AI's anthropomorphic structure form. Technological Forecasting and Social Change. 2025 Apr;213:124004.

38. Ferber D, El Nahhas OSM, Wölflein G, Wiest IC, Clusmann J, Leßmann ME, et al. Development and validation of an autonomous artificial intelligence agent for clinical decision-making in oncology. Nat Cancer. 2025 Jun 6;6(8):1337--49.

39. Marshall A. Drugmakers share data to feed voracious foundation models. Nat Biotechnol. 2025 Nov;43(11):1743--6.

40. Hor T (Shoon C, Fong L, Wynne K, Verhoeven B. Easing the cognitive load of general practitioners: AI design principles for future-ready healthcare. Technovation. 2025 Apr;142:103208.

41. Ahadian P, Xu W, Liu D, Guan Q. Ethics of trustworthy AI in healthcare: Challenges, principles, and practical pathways. Neurocomputing. 2026 Jan;661:131942.

42. Croxford E, Gao Y, First E, Pellegrino N, Schnier M, Caskey J, et al. Evaluating clinical AI summaries with large language models as judges. npj Digit Med. 2025 Nov 5;8(1):640.

43. Gaber F, Shaik M, Allega F, Bilecz AJ, Busch F, Goon K, et al. Evaluating large language model workflows in clinical decision support for triage and referral and diagnosis. npj Digit Med. 2025 May 9;8(1):263.

44. Mohammadi M, Li Y, Lo J, Yip W. Evaluation and Benchmarking of LLM Agents: A Survey. In Arxiv; 2025. p. 6129--39. Available from: http://arxiv.org/abs/2507.21504

45. Nair S. Explainable AI in Gxp Validation: Balancing Automation, Traceability, And Regulatory Trust in The Pharmaceutical Industry. Clinical Medicine and Health Research Journal. 2025

46. For trustworthy AI, keep the human in the loop. Nat Med. 2025 Oct;31(10):3207--3207.

47. Madrid-García A, Benavent D, Merino-Barbancho B. From chat to act: large language model agents and agentic AI as the next frontier of AI in rheumatology. EULAR Rheumatology Open. 2025 Oct;1(3):147--56.

48. Mahajan A, Powell D. Generalist medical AI reimbursement challenges and opportunities. npj Digit Med. 2025 Feb 26;8(1):125.

49. Harrer S, Rane RV, Speight RE. Generative AI agents are transforming biology research: high resolution functional genome annotation for multiscale understanding of life. eBioMedicine. 2024 Nov;109:105446.

50. Campos HDO, Wolfe D, Luan H, Sim I. Generative AI as Third Agent: Large Language Models and the Transformation of the Clinician-Patient Relationship. J Particip Med. 2025 Aug 11;17:e68146--e68146.

51. Reddy S. Generative AI in healthcare: an implementation science informed translational path on application, integration and governance. Implementation Sci. 2024 Mar 15;19(1):27.

52. Baig MM, Hobson C, GholamHosseini H, Ullah E, Afifi S. Generative AI in Improving Personalized Patient Care Plans: Opportunities and Barriers Towards Its Wider Adoption. Applied Sciences. 2024 Nov 25;14(23):10899.

53. Teo ZL, Thirunavukarasu AJ, Elangovan K, Cheng H, Moova P, Soetikno B, et al. Generative artificial intelligence in medicine. Nat Med. 2025 Oct;31(10):3270--82.

54. Blease C, Garcia Sanchez C, Locher C, McMillan B, Gaab J, Torous J. Generative Artificial Intelligence in Primary Care: Qualitative Study of UK General Practitioners' Views. J Med Internet Res. 2025 Aug 6;27:e74428--e74428.

55. Bhuyan SS, Sateesh V, Mukul N, Galvankar A, Mahmood A, Nauman M, et al. Generative Artificial Intelligence Use in Healthcare: Opportunities for Clinical Excellence and Administrative Efficiency. J Med Syst. 2025 Jan 16;49(1):10.

56. International Medical Device Regulators Forum. Good machine learning practice for medical device development: Guiding principles. 2025;

57. Glaser M, Littlebury R. Governance of artificial intelligence and machine learning in pharmacovigilance: what works today and what more is needed? Therapeutic Advances in Drug Safety. 2024 Jan;15:20420986241293303.

58. Lee Y, Ferber D, Rood JE, Regev A, Kather JN. How AI agents will change cancer research and oncology. Nat Cancer. 2024 Dec 17;5(12):1765--7.

59. Van Leersum CM, Maathuis C. Human centred explainable AI decision-making in healthcare. Journal of Responsible Technology. 2025 Mar;21:100108.

60. Romero-Obon M, Rouaz-El-Hajoui K, Sancho-Ochoa V, Vargas R, Pérez-Lozano P, Suñé-Pou M, et al. Human-in-the-Loop AI Use in Ongoing Process Verification in the Pharmaceutical Industry. Information. 2025 Dec 6;16(12):1082.

61. Merdin-Uygur E, Ozturkcan S, Özbilgin MF, Yılmaz F, İnce Ö. Human--robot collaboration in surgery at the nexus of knowledge, agency, and ownership. Sci Rep. 2025 Jul 2;15(1):23642.

62. Ostermann M, Freyer O, Verhees FG, Kather JN, Gilbert S. If a therapy bot walks like a duck and talks like a duck then it is a medically regulated duck. npj Digit Med. 2025 Dec 5;8(1):741.

63. Dennstädt F, Hastings J, Putora PM, Schmerder M, Cihoric N. Implementing large language models in healthcare while balancing control, collaboration, costs and security. npj Digit Med. 2025 Mar 6;8(1):143.

64. Bidenko NV, Stuchynska NV, Palamarchuk YV, Matviienko MM. Integrating artificial intelligence in healthcare practice: challenges and future prospects. Wiadomości Lekarskie. 2005 May 30;(5):1199--205.

65. Ong JCL, Ning Y, Collins GS, Bitterman DS, Beecy AN, Chang RT, et al. International partnership for governing generative artificial intelligence models in medicine. Nat Med. 2025 Sep;31(9):2836--9.

66. Goodell AJ, Chu SN, Rouholiman D, Chu LF. Large language model agents can use tools to perform clinical calculations. npj Digit Med. 2025 Mar 17;8(1):163.

67. Wiest IC, Bhat M, Clusmann J, Schneider CV, Jiang X, Kather JN. Large language models for clinical decision support in gastroenterology and hepatology. Nat Rev Gastroenterol Hepatol. 2025 Nov;22(11):773--87.

68. Kim DS, Eltahir A, Mantena S, Doijad R, Matabang C, Rodriguez F. Leveraging large language models to bridge the digital divide in cardiovascular health research. npj Cardiovasc Health. 2025 Dec 2;2(1):60.

69. Kim J, Podlasek A, Shidara K, Liu F, Alaa A, Bernardo D. Limitations of large language models in clinical problem-solving arising from inflexible reasoning. Sci Rep. 2025 Nov 11;15(1):39426.

70. Qiu J, Lam K, Li G, Acharya A, Wong TY, Darzi A, et al. LLM-based agentic systems in medicine and healthcare. Nat Mach Intell. 2024 Dec 5;6(12):1418--20.

71. Jiang Y, Black KC, Geng G, Park D, Zou J, Ng AY, et al. MedAgentBench: A Virtual EHR Environment to Benchmark Medical LLM Agents. NEJM AI. 2025 Aug 28;2(9). Available from: https://ai.nejm.org/doi/10.1056/AIdbp2500144

72. Wind S, Sopa J, Truhn D, Lotfinia M, Nguyen TT, Bressem K, et al. Multi-step retrieval and reasoning improves radiology question answering with large language models. npj Digit Med. 2025 Dec 22;8(1):790.

73. Borkowski A. Multiagent AI Systems in Health Care: Envisioning Next-Generation Intelligence. Federal Practitioner. 2025 May;42(5). Available from: https://www.mdedge.com/fedprac/article/272525/health-policy/multiagent-ai-systems-health-care-envisioning-next-generation-intelligence

74. Karunanayake N. Next-generation agentic AI for transforming healthcare. Informatics and Health. 2025 Sep;2(2):73--83.

75. Croatti A, Gabellini M, Montagna S, Ricci A. On the Integration of Agents and Digital Twins in Healthcare. J Med Syst. 2020 Sep;44(9):161.

76. Freyer O, Jayabalan S, Kather JN, Gilbert S. Overcoming regulatory barriers to the implementation of AI agents in healthcare. Nat Med. 2025 Oct;31(10):3239--43.

77. Favini N, Batlivala N, Mijic M, Daneshjou R. Policy brief: AI-first Medicaid: how CMS can build a smarter safety net with Precision Benefits. npj Digit Med. 2025 Nov 28;8(1):734.

78. Khalil RA, Ahmad K, Ali H. Redefining Elderly Care with Agentic AI: Challenges and Opportunities. arXiv; 2025. Available from: http://arxiv.org/abs/2507.14912

79. European Medicines Agency. Reflection paper on the use of Artificial Intelligence (AI) in the medicinal product lifecycle. 2024;

80. Niazi SK. Regulatory Perspectives for AI/ML Implementation in Pharmaceutical GMP Environments. Pharmaceuticals. 2025 Jun 16;18(6):901.

81. Davidovic J. Rethinking human roles in AI warfare. Nat Mach Intell. 2025 Oct 13;7(10):1593--5.

82. Shao E, Wang Y, Qian Y, Pan Z, Liu H, Wang D. SciSciGPT: advancing human--AI collaboration in the science of science. Nat Comput Sci. 2025 Dec 9; Available from: https://www.nature.com/articles/s43588-025-00906-6

83. Li D, Jiang N, Huang K, Tu R, Ouyang S, Yu H, et al. Streamlining evidence based clinical recommendations with large language models. npj Digit Med. 2025 Dec 22;8(1):793.

84. Lakhan SE. The Agentic Era: Why Biopharma Must Embrace Artificial Intelligence That Acts, Not Just Informs. Cureus. 2025 May 3; Available from: https://www.cureus.com/articles/363846-the-agentic-era-why-biopharma-must-embrace-artificial-intelligence-that-acts-not-just-informs

85. Albaroudi E, Mansouri T, Alameer A. The Intersection of Generative AI and Healthcare: Addressing Challenges to Enhance Patient Care. In: 2024 Seventh International Women in Data Science Conference at Prince Sultan University (WiDS PSU). Riyadh, Saudi Arabia: IEEE; 2024. p. 134--40. Available from: https://ieeexplore.ieee.org/document/10571863/

86. Reddy CKK, Joseph S, Joshi H, Doss S, Ouaissa M, editors. The Power of Agentic AI: Redefining Human Life and Decision-Making: In Industry 6.0. Cham: Springer Nature Switzerland; 2025. (Sustainable Artificial Intelligence-Powered Applications). Available from: https://link.springer.com/10.1007/978-3-031-89424-4

87. Zou J, Topol EJ. The rise of agentic AI teammates in medicine. The Lancet. 2025 Feb;405(10477):457.

88. Collaco BG, Haider SA, Prabha S, Gomez-Cabello CA, Genovese A, Wood NG, Bagaria SP, Gopala N, Tao C, Forte AJ. The role of agentic artificial intelligence in healthcare: a scoping review. npj Digital Medicine. 2026 Mar 14;9(1):345.

89. Gallifant J, Afshar M, Ameen S, Aphinyanaphongs Y, Chen S, Cacciamani G, et al. The TRIPOD-LLM reporting guideline for studies using large language models. Nat Med. 2025 Jan;31(1):60--9.

90. Huang L, Koutra D, Kulkarni A, Prioleau T, Wu Q, Yan Y, et al. Towards Agentic AI for Science: Hypothesis Generation, Comprehension, Quantification, and Validation. In: Companion Proceedings of the ACM on Web Conference 2025. Sydney NSW Australia: ACM; 2025. p. 1639--42. Available from: https://dl.acm.org/doi/10.1145/3701716.3717754

91. Xin H, Kitchin JR, Kulik HJ. Towards agentic science for advancing scientific discovery. Nat Mach Intell. 2025 Sep 10;7(9):1373--5.

92. Mohsin Khan M, Shah N, Shaikh N, Thabet A, Alrabayah T, Belkhair S. Towards secure and trusted AI in healthcare: A systematic review of emerging innovations and ethical challenges. International Journal of Medical Informatics. 2025 Mar;195:105780.

93. Ismail M, Barth H, Holmén M, Petersson L, Irgang L. Transforming towards AI-augmented Healthcare: Experiences of physicians in Sweden. Technovation. 2025 Dec;148:103333.

94. Li H, Pan W, Rajendran S, Zang C, Wang F. TrialGenie: Empowering Clinical Trial Design with Agentic Intelligence and Real World Data. MedRxiv. 2025

95. Hassan R, Nguyen N, Finserås SR, Adde L, Strümke I, Støen R. Unlocking the black box: Enhancing human-AI collaboration in high-stakes healthcare scenarios through explainable AI. Technological Forecasting and Social Change. 2025 Oct;219:124265.

96. Lagarde W, Glassbox Bio. VECTR: Towards a Reliability Framework for Agentic AI in Drug Development. Zenodo Preprint. 2026 Jan 2; Available from: https://zenodo.org/doi/10.5281/zenodo.18122849

97. Gabriel I, Keeling G, Manzini A, Evans J. We need a new ethics for a world of AI agents. Nature. 2025;

98. Kirk HR, Gabriel I, Summerfield C, Vidgen B, Hale SA. Why human--AI relationships need socioaffective alignment. Humanit Soc Sci Commun. 2025 May 28;12(1):728.

99. Yue L, Xing S, Chen J, Fu T. ClinicalAgent: Clinical Trial Multi-Agent System with Large Language Model-based Reasoning. arXiv; 2024 . Available from: http://arxiv.org/abs/2404.14777

100. FDA. FDA Expands Artificial Intelligence Capabilities with Agentic AI Deployment. 2025. Available from: https://www.fda.gov/news-events/press-announcements/fda-expands-artificial-intelligence-capabilities-agentic-ai-deployment

101. EU AI Act. High-level summary of the AI Act. 2024. Available from: https://artificialintelligenceact.eu/high-level-summary/

102. Bashir A, han TA, Shamszaman ZU. Many-to-One Adversarial Consensus: Exposing Multi-Agent Collusion Risks in AI-Based Healthcare. arXiv; 2025 . Available from: http://arxiv.org/abs/2512.03097

103. Kim Y, Jeong H, Chen S, Li SS, Park C, Lu M, et al. Medical Hallucination in Foundation Models and Their Impact on Healthcare. 2025. Available from: http://medrxiv.org/lookup/doi/10.1101/2025.02.28.25323115

104. Xu S, Yan Z, Dai C, Wu F. MEGA-RAG: a retrieval-augmented generation framework with multi-evidence guided answer refinement for mitigating hallucinations of LLMs in public health. Front Public Health. 2025 Oct 8;13:1635381.

105. Han X, Gao X, Qu X, Yu Z. Multi-Agent Medical Decision Consensus Matrix System: An Intelligent Collaborative Framework for Oncology MDT Consultations. arXiv; 2025 . Available from: http://arxiv.org/abs/2512.14321

106. FDA. Predetermined Change Control Plans for Machine Learning-Enabled Medical Devices: Guiding Principles. 2025. Available from: https://www.fda.gov/medical-devices/software-medical-device-samd/predetermined-change-control-plans-machine-learning-enabled-medical-devices-guiding-principles

107. Carvalho E, Mascarenhas M, Pinheiro F, Correia R, Balseiro S, Barbosa G, et al. Predetermined Change Control Plans: Guiding Principles for Advancing Safe, Effective, and High-Quality AI-ML Technologies. JMIR AI. 2025 Oct 31;4:e76854--e76854.

108. Meimandi KJ, Aránguiz-Dias G, Kim GR, Saadeddin L, Griffith A, Kochenderfer MJ. The Measurement Imbalance in Agentic AI Evaluation Undermines Industry Productivity Claims. arXiv; 2025 . Available from: http://arxiv.org/abs/2506.02064

109. Xu, T., Zhang, D., Mitra, K. and Hruschka, E., 2025. Verification-aware planning for multi-agent systems. arXiv preprint arXiv:2510.17109.

110. FDA. Computer Software Assurance for Production and Quality System Software; 2025. https://www.fda.gov/media/188844/download

111. Plale, B., Karthikeyan, N., Gamage, I., Stubbs, J. and Withana, S., 2025. AI/ML Model Cards in Edge AI Cyberinfrastructure: towards Agentic AI. arXiv preprint arXiv:2511.21661. https://arxiv.org/abs/2511.21661

112. Enqvist, L., 2023. `Human oversight'in the EU artificial intelligence act: what, when and by whom?. Law, Innovation and Technology, 15(2), pp.508-535.

113. EU AI Act: Article 14: Human Oversight. https://artificialintelligenceact.eu/article/14/
\endgroup

\clearpage
\begingroup
\footnotesize
\setlength{\tabcolsep}{3pt}
\renewcommand{\arraystretch}{1.18}

\newcommand{\coreqMethods}{Methods, Subject matter expert interviews, pp.~20--21}
\newcommand{\coreqMethodsLater}{Methods, Subject matter expert interviews, p.~21}
\newcommand{\coreqResults}{Results, Expert perspectives, p.~6}
\newcommand{\coreqSupplement}{Supplementary Note 2, p.~28}

\begin{center}
\textbf{VITAL: A framework for validating agentic AI in pharmaceutical drug development and clinical trial operations}
\end{center}

\section*{COREQ (COnsolidated criteria for REporting Qualitative research) Checklist}

\noindent\textit{Developed from:} Tong A, Sainsbury P, Craig J. Consolidated criteria for reporting qualitative research (COREQ): a 32-item checklist for interviews and focus groups. \textit{Int J Qual Health Care}. 2007;19(6):349--357.

\begin{longtable}{@{}
  >{\raggedright\arraybackslash}p{0.19\linewidth}
  >{\centering\arraybackslash}p{0.055\linewidth}
  >{\raggedright\arraybackslash}p{0.38\linewidth}
  >{\raggedright\arraybackslash}p{0.30\linewidth}@{}}

\toprule
\textbf{Topic} & \textbf{Item No.} &
\textbf{Guide Questions/Description} &
\textbf{Reported on / Page No.} \\
\midrule
\endfirsthead

\toprule
\textbf{Topic} & \textbf{Item No.} &
\textbf{Guide Questions/Description} &
\textbf{Reported on / Page No.} \\
\midrule
\endhead

\bottomrule
\endfoot

\bottomrule
\endlastfoot

\addlinespace[3pt]
\multicolumn{4}{@{}l}{\textbf{Domain 1: Research team and reflexivity}} \\
\addlinespace[2pt]
\multicolumn{4}{@{}l}{\textit{Personal characteristics}} \\

Interviewer/facilitator & 1 &
Which author/s conducted the interview or focus group? &
\coreqMethods \\

Credentials & 2 &
What were the researcher's credentials? E.g. PhD, MD &
\coreqMethods \\

Occupation & 3 &
What was their occupation at the time of the study? &
\coreqMethods \\

Gender & 4 &
Was the researcher male or female? &
Not disclosed \\

Experience and training & 5 &
What experience or training did the researcher have? &
\coreqMethods \\

\multicolumn{4}{@{}l}{\textit{Relationship with participants}} \\

Relationship established & 6 &
Was a relationship established prior to study commencement? &
\coreqMethods \\

Participant knowledge of the interviewer & 7 &
What did the participants know about the researcher? e.g. personal goals, reasons for doing the research &
\coreqMethods \\

Interviewer characteristics & 8 &
What characteristics were reported about the interviewer/facilitator? e.g. Bias, assumptions, reasons and interests in the research topic &
\coreqMethods \\

\addlinespace[3pt]
\multicolumn{4}{@{}l}{\textbf{Domain 2: Study design}} \\
\addlinespace[2pt]
\multicolumn{4}{@{}l}{\textit{Theoretical framework}} \\

Methodological orientation and Theory & 9 &
What methodological orientation was stated to underpin the study? e.g. grounded theory, discourse analysis, ethnography, phenomenology, content analysis &
\coreqMethods \\

\multicolumn{4}{@{}l}{\textit{Participant selection}} \\

Sampling & 10 &
How were participants selected? e.g. purposive, convenience, consecutive, snowball &
\coreqMethods \\

Method of approach & 11 &
How were participants approached? e.g. face-to-face, telephone, mail, email &
\coreqMethods \\

Sample size & 12 &
How many participants were in the study? &
\coreqMethods \\

Non-participation & 13 &
How many people refused to participate or dropped out? Reasons? &
\coreqMethods \\

\multicolumn{4}{@{}l}{\textit{Setting}} \\

Setting of data collection & 14 &
Where was the data collected? e.g. home, clinic, workplace &
\coreqMethods \\

Presence of non-participants & 15 &
Was anyone else present besides the participants and researchers? &
\coreqMethods \\

Description of sample & 16 &
What are the important characteristics of the sample? e.g. demographic data, date &
\coreqMethods; \coreqResults \\

\multicolumn{4}{@{}l}{\textit{Data collection}} \\

Interview guide & 17 &
Were questions, prompts, guides provided by the authors? Was it pilot tested? &
\coreqMethodsLater; \coreqSupplement \\

Repeat interviews & 18 &
Were repeat interviews carried out? If yes, how many? &
\coreqMethods \\

Audio/visual recording & 19 &
Did the research use audio or visual recording to collect the data? &
\coreqMethods \\

Field notes & 20 &
Were field notes made during and/or after the interview or focus group? &
\coreqMethods \\

Duration & 21 &
What was the duration of the interviews or focus group? &
\coreqMethods \\

Data saturation & 22 &
Was data saturation discussed? &
\coreqMethods \\

Transcripts returned & 23 &
Were transcripts returned to participants for comment and/or correction? &
\coreqMethods \\

\addlinespace[3pt]
\multicolumn{4}{@{}l}{\textbf{Domain 3: analysis and findings}} \\
\addlinespace[2pt]
\multicolumn{4}{@{}l}{\textit{Data analysis}} \\

Number of data coders & 24 &
How many data coders coded the data? &
\coreqMethodsLater \\

Description of the coding tree & 25 &
Did authors provide a description of the coding tree? &
\coreqMethodsLater \\

Derivation of themes & 26 &
Were themes identified in advance or derived from the data? &
\coreqMethodsLater \\

Software & 27 &
What software, if applicable, was used to manage the data? &
\coreqMethodsLater \\

Participant checking & 28 &
Did participants provide feedback on the findings? &
\coreqMethodsLater \\

\multicolumn{4}{@{}l}{\textit{Reporting}} \\

Quotations presented & 29 &
Were participant quotations presented to illustrate the themes/findings? Was each quotation identified? e.g. participant number &
\coreqMethodsLater \\

Data and findings consistent & 30 &
Was there consistency between the data presented and the findings? &
\coreqResults \\

Clarity of major themes & 31 &
Were major themes clearly presented in the findings? &
\coreqResults \\

Clarity of minor themes & 32 &
Is there a description of diverse cases or discussion of minor themes? &
\coreqResults \\

\end{longtable}
\endgroup

\end{document}